\documentclass[preprint,5p,times,twocolumn]{elsarticle}

\usepackage{amssymb}
\usepackage{amsmath}

\usepackage{booktabs}
\usepackage{url}
\usepackage{algorithm}
\usepackage{algpseudocode}
\usepackage{pdflscape}
\usepackage{wrapfig}
\usepackage{subcaption}
\usepackage{csquotes}
\usepackage{geometry}
\usepackage{hyperref}

\usepackage{lineno}

\journal{Pattern Recognition Letters} 

\begin{document}

\begin{frontmatter}



\title{Ctrl-Crash: Controllable Diffusion for Realistic Car Crashes}


\affiliation[1]{organization={Mila},
            city={Montreal},
            state={Quebec},
            country={Canada}}

\affiliation[2]{organization={Polytechnique Montréal},
            city={Montreal},
            state={Quebec},
            country={Canada}}

\affiliation[3]{organization={Université de Montréal},
            city={Montreal},
            state={Quebec},
            country={Canada}}

\affiliation[4]{organization={McGill University},
            city={Montreal},
            state={Quebec},
            country={Canada}}

\affiliation[5]{organization={CIFAR AI Chair},
            city={Toronto}, 
            country={Canada}}

\affiliation[6]{organization={Samsung SAIL Montréal},
            city={Montreal},
            state={Quebec},
            country={Canada}}


\author[1,2]{Anthony Gosselin}
\corref{cor1}
\ead{anthony.gosselin@mila.quebec}

\author[1,3]{Ge Ya Luo}
\ead{xugeya@mila.quebec}

\author[1]{Luis Lara}
\ead{luis.lara@mila.quebec}

\author[1]{Florian Golemo}
\ead{golemofl@mila.quebec}

\author[1,4,5]{Derek Nowrouzezahrai}
\ead{derek@mila.quebec}

\author[1,3,5]{Liam Paull}
\ead{paulll@iro.umontreal.ca}

\author[6]{Alexia Jolicoeur-Martineau}
\ead{alexia.jolicoeur-martineau@mail.mcgill.ca}

\author[1,2,5]{Christopher Pal}
\ead{christopher.pal@mila.quebec}

\cortext[cor1]{Corresponding author}

\begin{abstract}
Video diffusion techniques have advanced significantly in recent years; however, they struggle to generate realistic imagery of car crashes due to the scarcity of accident events in most driving datasets. 
Improving traffic safety requires realistic and controllable accident simulations. To tackle the problem, we propose Ctrl-Crash, a controllable car crash video generation model that conditions on signals such as bounding boxes, crash types, and an initial image frame.  
Our approach enables counterfactual scenario generation where minor variations in input can lead to dramatically different crash outcomes. To support fine-grained control at inference time, we leverage classifier-free guidance with independently tunable scales for each conditioning signal. 
Ctrl-Crash achieves state-of-the-art performance across quantitative video quality metrics (e.g., FVD and JEDi) and qualitative measurements based on a human-evaluation of physical realism and video quality compared to prior diffusion-based methods. \\\textbf{\small Project page: \url{https://anthonygosselin.github.io/Ctrl-Crash-ProjectPage/}}
\end{abstract}


\end{frontmatter}

\section{Introduction}
\begin{figure*}
    \centering
    \vspace{-10mm}
    \includegraphics[width=0.85\linewidth]{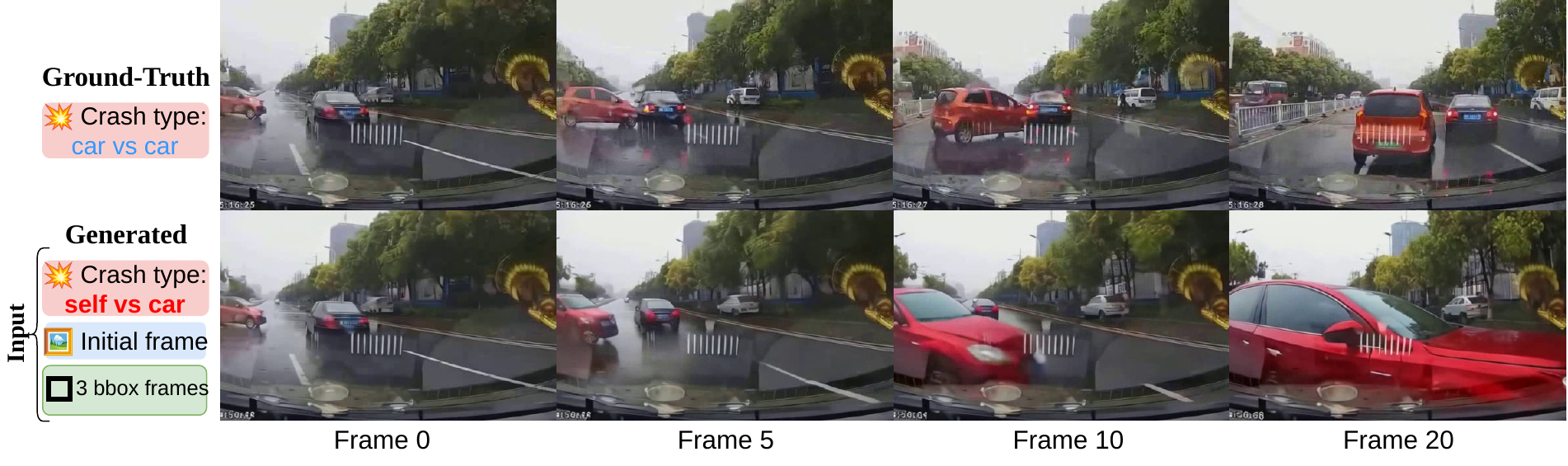}
    \vspace{-3mm}
    \caption{\textbf{Counterfactual Crash Generation}: this diagram demonstrates the ability of our model to generate counterfactual crashes while beginning from the identical initial frame. \emph{Top:} a ground truth accident between two  vehicles other than the ego-vehicle, where the red car hits the rear of the blue car and spins into the lane in front of the ego-vehicle. \emph{Bottom:} the model generates an alternative accident involving the ego-vehicle. In this alternative future the red car avoids the blue car but turns into the path of the ego-vehicle leading to the crash.}
    \label{fig:counter_red}
\end{figure*}

Autonomous vehicle (AV) systems must be rigorously tested in a wide range of driving situations—including rare and dangerous edge cases such as collisions to ensure safe deployment. Much of the current progress in perception, planning, and control for AVs has been driven by large-scale datasets collected in harmless, non-crash scenarios. However, realistic video data of car crashes remains extremely scarce, making it difficult to simulate, anticipate, or learn effectively from these critical events~\cite{wang2023deepaccident}. 

Prior work has largely approached the challenge of crash scenario modeling in two main ways. 

On one hand, physics-based rendering approaches use game engines or physics simulators to model accident dynamics, but often fall short on visual realism, require expensive rendering pipelines, and demand costly human effort for environment and asset creation~\cite{dosovitskiy17carla}. On the other hand, data-driven methods, such as generative models, rely on real-world footage, which is difficult to acquire in sufficient volume due to the infrequency and ethical complexity of crash events~\cite{holger2019nuscnes}. Moreover, most generative approaches focus on normal driving behavior, avoiding the complexity and unpredictability inherent in crash dynamics, where agent interactions, rare motion patterns, and semantic context all matter deeply~\cite{wu2024drivescape, luo2025ctrlv}.

To address this gap, we introduce \textbf{Ctrl-Crash}, a controllable video diffusion framework for generating realistic crash videos from a single initial frame. Our method operates with inputs and outputs in pixel space, as opposed to using computer graphics primitives and explicit models of physics. Our approach can generate video conditioned on an initial image frame, spatial control signals consisting of bounding box sequences of cars and pedestrians, and semantic intent signals encoded as discrete crash types, enabling the generation of diverse crash outcomes. Through these conditioning signals, we can direct the narrative of a crash, simulate plausible sequences of interactions, and explore counterfactual variants of a given scene, answering the following types of questions with high quality generated video: \textit{How might the scene evolve differently under a different agent trajectory or crash type?}

Ctrl-Crash builds on latent diffusion models \cite{rombach2022high} and classifier-free guidance \cite{ho2022cfg}, and we extend the latter to allow independently tunable guidance strengths for each control modality, making our system highly flexible at inference time. Our two-stage training procedure first finetunes a pretrained Stable Video Diffusion (SVD) \cite{blattmann2023svd} model on in-the-wild ego-view accident videos, then trains a ControlNet \cite{zhang2023controlnet} adapter to handle conditioning in order to direct the video generation. By leveraging this data-driven framework, our model can generate controllable crash videos that are visually realistic, semantically diverse, and behaviorally plausible. We see this work as a step toward not only improving the diversity and coverage of safety-critical AV testing, but also enabling counterfactual safety reasoning: the ability to simulate alternate outcomes from identical initial conditions, and better understand the causality of crashes.\\

\textbf{Our contributions:}
\begin{enumerate}
    \item We introduce Ctrl-Crash, a fully data-driven generative framework for realistic and controllable car crash video generation. Our method obtains state-of-the-art performance compared to prior diffusion-based car accident video generation methods in quantitative (e.g., FVD) and qualitative analysis (human-evaluation of physical realism and video quality).
    \item Our approach can generate plausible and diverse crash outcomes from the same initial frame and initial bounding boxes, enabling counterfactual video simulation for safety-critical reasoning. 
    \item We develop a data-processing pipeline to filter driving videos and extract bounding box trajectories of road users. Using it, we release a curated extension of the MM-AU dataset~\cite{fang2024abductive} (with bounding box annotations), as well as held-out test sets from Car Crash Dataset RUSSIA~\cite{sivoha2023carcrash} (with bounding box annotations) and BDD100k~\cite{yu2020bdd100k} (with existing boxes), along with tools to support research in crash simulation and controllable video generation. Code and dataset extensions are all made open-source.
\end{enumerate}

\section{Related Work}
\label{rel_work}
\textbf{Video Diffusion Models.}
Diffusion models~\cite{ho2020ddpm} have emerged as a powerful paradigm for generative modeling, particularly in the domain of image synthesis. They operate by learning to reverse a gradual noising process applied to data, generating high-fidelity samples through iterative denoising. Recent advances have extended these techniques to the video domain, where temporal consistency and spatial coherence are critical~\cite{ho2022vdm, blattmann2023svd}. 


Given the noise-level $t$, and task-specific conditions $c$, the diffusion loss is the following: 

\begin{equation}
    \mathcal{L} = \mathbb{E}_{x_0, t, c, \epsilon \sim \mathcal{N}(0, \mathbf{I}) }\Big[ \Vert \epsilon - \epsilon_\theta(x_{t}, t, c)) \Vert_{2}^{2}\Big],
    \label{eq:loss}
\end{equation}

where $x_t = \sqrt{\bar\alpha_t} x_0 + \sqrt{1-\bar\alpha_t}\epsilon$, $\epsilon \sim \mathcal{N}(0, \mathbf{I})$, $x_0$ is a training sample video,  $\bar\alpha_t$ at $t \in [1,T]$ controls the diffusion schedule, and $\epsilon_\theta$ is a function approximator intended to predict the noise $\epsilon$ from the noise-corrupted video $x_t$.

Latent video diffusion models (LVDMs)~\cite{he2023lvdm, blattmann2023svd} address the computational challenges of high-resolution video generation by operating in a compressed latent space. This enables the generation of long, high-quality video sequences from compact representations. Specifically, Stable Video Diffusion (SVD)~\cite{blattmann2023svd}, an LVDM variant, leverages a UNet-based denoiser trained on video latents conditioned on an initial frame, making it suitable for tasks like image-to-video generation and temporal extension.

Our work builds on this foundation by fine-tuning a pretrained SVD model on a large curated car crash dataset and giving it additional controllability mechanisms to make it well-suited for generating complex driving scenes and crash dynamics.

\textbf{Controllable Generative Models.}
Recent progress in generative modeling has emphasized not only fidelity, but also controllability — the ability to guide outputs through structured input signals. In image generation, this includes control via text prompts, sketches, bounding boxes, semantic maps, and more. In diffusion models, classifier-free guidance (CFG)~\cite{ho2022cfg} and ControlNet~\cite{zhang2023controlnet} have been introduced as effective methods to allow adaptable conditioning while preserving high-quality generation.

CFG is a widely-used technique in diffusion models to improve conditional generation by combining conditional and unconditional predictions, scaled to control how strongly the model follows the conditioning input. It involves jointly training the diffusion model for conditional and unconditional denoising by randomly setting the conditioning to a null value $c=\emptyset$ during training. During inference, the denoising prediction is interpolated between a conditional noise estimate $\epsilon_\theta(\mathbf{x}_t, c)$ and an unconditional noise estimate $\epsilon_\theta(\mathbf{x}_t, \emptyset)$, scaled by a guidance factor $\gamma$ to obtain the modified score estimate:
\begin{equation}
\label{eq:cfg_eqn}
    \hat{\epsilon}_\theta(\mathbf{x}_t, c) = 
    \epsilon_\theta(\mathbf{x}_t, \emptyset) + \gamma \cdot \left( \epsilon_\theta(\mathbf{x}_t, c) - \epsilon_\theta(\mathbf{x}_t, \emptyset) \right).
\end{equation}
Vision models, such as InstructPix2Pix~\cite{pix2pix}, use textual or mixed-mode conditioning with CFG to manipulate generation intent and content mid-sampling. ControlNet~\cite{zhang2023controlnet} introduced an effective approach for injecting spatial control into diffusion models by adding a parallel, trainable network that processes conditioning signals and modulates the main denoising backbone. It works by branching off from intermediate layers in the main U-Net and processing a control input in parallel. Its outputs are then added to the original U-Net features before the denoising step, effectively steering the generation without retraining the base model.

Our method advances prior research by combining semantic control (crash type) and spatial control (bounding box) through CFG and via the ControlNet adapter. This integration enables both precise descriptive control and generative reasoning about critical driving outcomes.

\textbf{Car Crash Simulation and AV Safety.}
Video diffusion models offer a compelling solution for car crash simulation, as they can simulate both the visual realism and behavioral dynamics of complex driving scenes with very little user effort. Recent car video generation models \cite{zhao2025drivedreamer, luo2025ctrlv, agarwal2025cosmos} focus on structured driving video generation using control signals like bounding boxes or text prompts. These models generate temporally consistent driving scenes and support the synthesis of structured traffic interactions. However, they are typically limited to non-crash scenarios or coarse control.

Physics-based driving simulators \cite{dosovitskiy17carla} are useful tools for evaluating autonomous vehicle safety \cite{rowe2024ctrlsim, dupuy2001generating, wang2024deepaccident}. Due to the scarcity and ethical challenges of real-world crash data, there is growing interest in generating synthetic safety-critical scenarios \cite{wang2024deepaccident, huang2025safety}. These simulations provide control and physical realism but often lack visual authenticity and broad applicability. In contrast, our work focuses on rare, adversarial cases that challenge perception and planning systems in real-world scenarios. To address the data challenge, we propose a method for processing car crash videos.

While most prior work focus on general driving scenes, notable examples of prior work tackling realistic car crash generation using real videos include:

DrivingGen \cite{guo2024drivinggen}, which generates crash videos from textual accident reports, AVD2 \cite{li2025avd2} and OAVD \cite{fang2024abductive} which also generate videos of car crashes, but their main focus is car crash video description rather than generation. 
These methods are designed to stress-test AV policies and enhance safety coverage beyond what real-world datasets offer. 

Our work builds on this vision by proposing a controllable video diffusion model that can generate crash outcomes directly from initial conditions and desired outcomes. Unlike physics simulators, Ctrl-Crash supports semantic control (e.g., crash type) and spatial trajectory specification (e.g., bounding boxes), enabling targeted stress testing of vision-based AV stacks. Moreover, our method supports counterfactual safety reasoning, allowing one to explore how small changes in agent behavior or intent could lead to dramatically different outcomes—a critical capability for understanding near-miss events and decision boundary failures. Importantly, our method has much higher visual and motion fidelity than prior works. 

\section{Our Method: Ctrl-Crash} \label{sec:method}
\begin{figure*}[t]
  \centering
  \vspace{-12mm} 
  \includegraphics[width=0.91\linewidth]{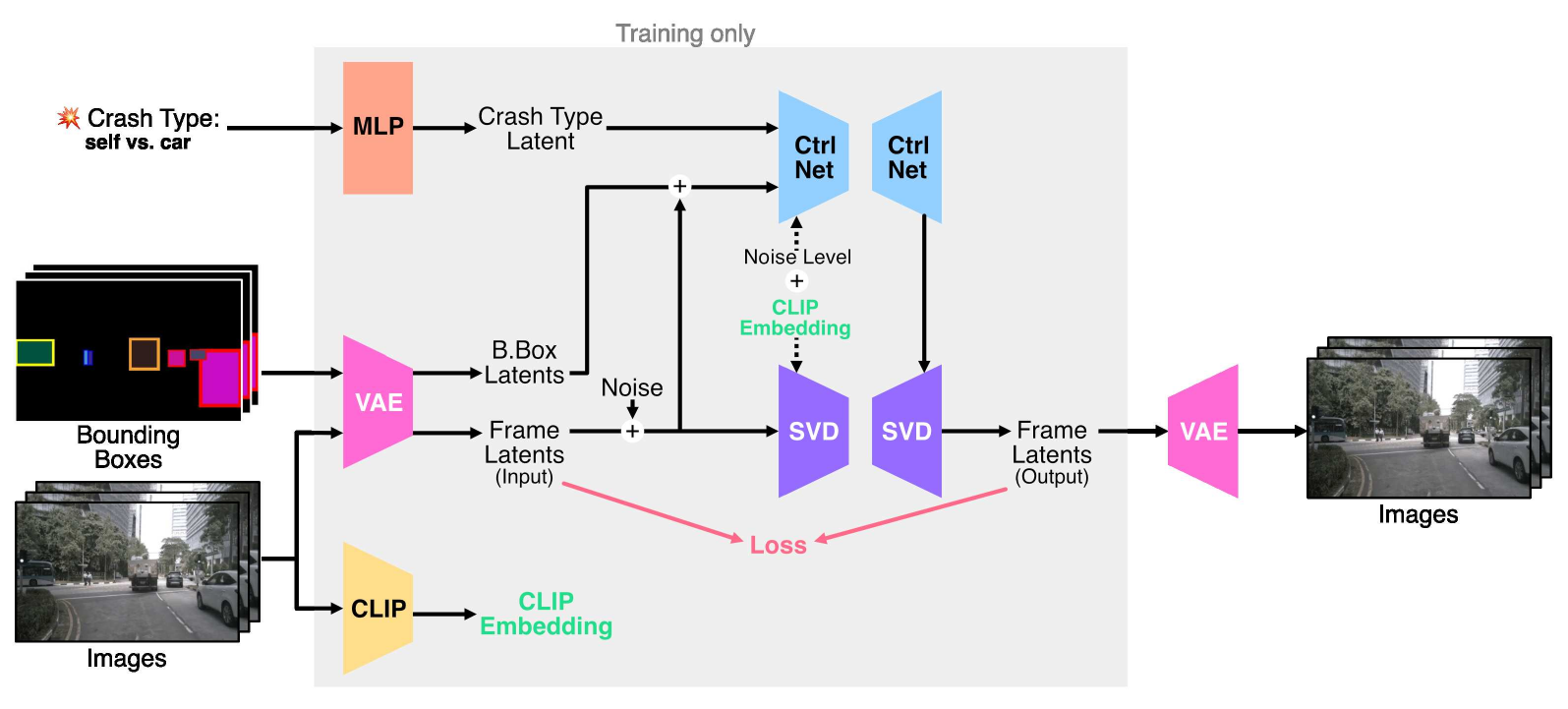}
  \vspace{-3mm}
  \caption{\textbf{Ctrl-Crash architecture}: Ctrl-Crash treats Bounding Boxes (BBs) as images. Both BBs and images frames are put through a VAE image encoder. The crash type and BB embeddings are fed to ControlNet. The images embedding after adding noise ($x_t$), the noise level ($t$), and the ControlNet intermediate outputs ($c$) are fed to the Stable Video Diffusion (SVD) model to obtain the predicted noise $\epsilon_\theta(x_{t}, t, c)$.  CLIP embeddings \citep{radford2021clip} are computed by passing the first image for each video through a pretrained CLIP encoder. These CLIP embeddings are then given to the ControlNet and SVD models. The diffusion process is solved over multiple steps using classifier-free guidance in the latent space, and then decoded back to images using the VAE image decoder. See Section \ref{sec:method} for details.}
  \label{fig:2}
\end{figure*}

In this section, we present \textit{Ctrl-Crash}, our controllable video diffusion framework for generating crash scenarios from a single image. We describe the overall architecture \ref{method_overview}, data processing pipeline \ref{method_dataprep}, conditioning mechanisms \ref{method_cond_signals}, training strategy \ref{method_training}, and our extension of classifier-free guidance for fine-grained control \ref{method_cfg}.

\subsection{Overview} 
\label{method_overview}
We propose Ctrl-Crash (Figure \ref{fig:2}), a controllable video diffusion framework designed to generate realistic car crash scenarios from a single initial frame, guided by both spatial and semantic control signals. Ctrl-Crash builds on \mbox{Ctrl-V~\cite{luo2025ctrlv}}, a framework for generating videos from rendered bounding box trajectories, by extending its capabilities to crash-specific scenarios, offering richer control and greater flexibility. Specifically, we incorporate a new semantic control signal representing crash type and introduce a refined training procedure to handle partial and noisy conditioning.

Our method follows a two-stage training pipeline. In the first stage, we fine-tune the Stable Video Diffusion (SVD) model on the MM-AU~\cite{fang2024abductive} crash video dataset to improve its ability to synthesize dynamic and physically plausible driving and accident scenes. In the second stage, we train a ControlNet module to inject conditioning information in two forms: (1) bounding box sequences representing road user motion across time, and (2) discrete crash type labels encoding high-level semantic intent. To ensure generalization to incomplete or noisy control, we introduce a curriculum-based random masking strategy that progressively masks out parts of the control inputs during training. Masked bounding box frames are replaced by a learnable embedding that preserves scene plausibility. We further extend classifier-free guidance to support independent scaling of each control modality, enabling nuanced and flexible control at inference. Unconditional noise predictions are obtained from the pretrained base model for improved generation diversity and stability.

\begin{figure}[!htb]
  \centering
  \includegraphics[width=\linewidth]{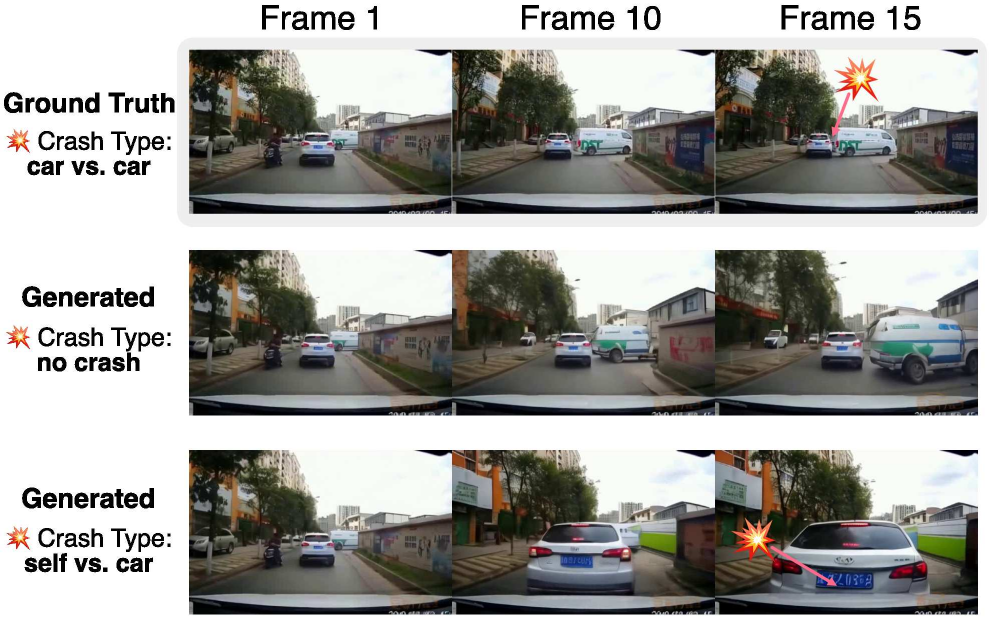}
  \caption{\textbf{Counterfactual Crash Generation}: this diagram demonstrates the ability of our model to generate counterfactual crashes (\textit{Middle}: no crash, \textit{Bottom}: ego/car crash) while beginning from the initial frame and 3 bounding-boxes frames of the real video (\textit{Top}: the real car crash).}
  \label{fig:counterfactual2}
\end{figure}

Ctrl-Crash supports three task settings, each enabled by varying the available control signals: 
(1)~\emph{Crash Reconstruction}: Given an initial image, full bounding box sequence, and a crash type, the model reconstructs a consistent video combining the visual context of the initial frame with agent motion derived from the bounding boxes. (2)~\emph{Crash Prediction}: Given the initial frame and either none or a few initial bounding box frames (e.g., 0–9), the model predicts the future motion of agents in a way that aligns with the target crash type. (3)~\emph{Crash Counterfactuals}: Extending the prediction task, this mode varies the crash type signal while keeping other inputs fixed, enabling the generation of multiple plausible outcomes for the same scene, supporting counterfactual safety reasoning (Figures \ref{fig:counter_red} and \ref{fig:counterfactual2}). We use the term ``counterfactual'' here in an informal sense, not to imply a formal causal model, but rather to describe alternate plausible outcomes from the same initial conditions. For example, Ctrl-Crash can be applied to non-crash datasets such as BDD100K, allowing the generation of realistic crash outcomes from otherwise uneventful driving scenes (see \ref{appendix:more_qual} for examples).

\subsection{Data Preparation}
\label{method_dataprep}
\label{sec_dataset}
Processing and preparing crash data is an essential element of our approach, enabling the construction of control signals from a diverse set of car crashes captured by dashboard cameras and sourced from online videos.

\textbf{Video Processing.} For training, we use the MM-AU dataset, a large-scale collection of dashcam crash videos collected from online sources. To ensure high quality we curate this dataset by applying a series of filtering steps. This includes automated detection and filtering of low resolution videos, shot change detection with PySceneDetect, and manual exclusion of scenes involving visible humans getting struck. We manually exclude scenes where visible humans are hit to avoid exposing the model to violent content and to prevent it from learning to depict human injury, reducing the risk of harmful or inappropriate generations post-deployment. Additionally, we standardize clips to 25-frame segments at 6 fps and $512 \times 320$ resolution. After the filtering steps, we retain approximately 7,500 videos from the original 11,727 videos. We split these videos into a training set and a held-out test set by randomly sampling by accident type categories with a 90/10 ratio. For more details on the precise processing steps and the dataset, please refer to \ref{appendix:video_proc} and \ref{appendix:datasets}.

\textbf{Bounding Box Extraction Pipeline.} To obtain reliable bounding box annotations for all road users, we design a hybrid pipeline that combines detection and segmentation models. For detection, we use YOLOv8~\cite{yolo2023} for frame-by-frame object detection. YOLO provides class-specific bounding boxes with high confidence and speed. For tracking, we use SAM2~\cite{ravi2024sam2} to produce instance-level masks and reliable tracking particularly when objects get occluded or deformed, common in crash videos.
This combined approach yields temporally aligned bounding boxes across all video frames. Importantly, it supports agents that enter or exit the scene dynamically—critical for realistic dynamic driving scenarios.
A full breakdown of the bounding-box annotation pipeline is provided in \ref{appendix:bbox_gen}. We collect bounding-box annotations for all road users for the MM-AU dataset and the RUSSIA Car Crash Dataset~\cite{sivoha2023carcrash}. These annotations along with the processed video frames will be made publicly available on our project page.

\subsection{Conditioning Signals} 
\label{method_cond_signals}
Ctrl-Crash generates videos from three complementary conditioning signals that guide both the visual realism and semantic plausibility of the crash scenario (as illustrated in Figure~\ref{fig:2}):

\textbf{Initial Frame (Scene Context):}
A single starting image grounds the generation process, providing the model with the appearance, layout, and environment of the driving scene before the crash occurs. It is encoded using a pretrained visual encoder (VAE) and fed into the main diffusion model as context.

\textbf{Bounding-Box Trajectories (Spatial Control):}
The motion of road users is guided by sequences of 2D bounding boxes, which encode object positions, dimensions, unique track ID (via fill color), and classes (via border color) over time. These control frames are encoded using the same pretrained VAE as the initial image and processed by a ControlNet and injected into the denoising process to allow the model to follow specific trajectories and react to interactions in a physically plausible way. See \ref{appendix:bbox_viz} for a visualization.

\textbf{Crash Type Label (Semantic Control):}
Crash types are represented by a discrete class label from five categories: (0) none, (1) ego-only, (2) ego/vehicle, (3) vehicle-only, and (4) vehicle/vehicle. These indicate which agents are involved in the crash, with, for example, “vehicle/vehicle” describing a crash between two non-ego agents (\textit{ego} being the vehicle from which we observe the scene from). The crash type index is embedded and projected into the cross-attention layers of the ControlNet encoder, allowing the model to generate outcomes consistent with high-level semantic intent.

\subsection{Training Pipeline}
\label{method_training}

Our training strategy is divided into two phases. In the first stage, we fine-tune the SVD model on crash-heavy video data from our preprocessed version of the MM-AU dataset (see Section \ref{sec_dataset}). The setup is an image-to-video task with an MSE loss in latent space (Equation~\ref{eq:loss}). This improves the model’s ability to generate plausible driving videos, including crashes, though without explicit controllability.
In the second stage, we freeze the fine-tuned base model and train a ControlNet adapter module with the same objective to introduce conditional controllability. This stage allows us to direct the generation using spatial and semantic control signals—namely, bounding box frames and crash type identifiers. Additional training and implementation details are provided in \ref{appendix:training} for reproducibility.

\textbf{Conditioning Masking.} To improve robustness and support flexible guidance at inference, we apply randomized masking of control signals during training. For bounding boxes, we use a temporal dropout strategy: a random timestep $k$ is sampled, and all subsequent frames are replaced with a \textit{learnable null embedding} to avoid misinterpreting missing input as the absence of agents. A curriculum schedule gradually increases the masking ratio to encourage generalization to partial control. For semantic inputs (initial image and crash type), we apply joint masking with a fixed probability schedule: with 10\% chance only the image is masked, 10\% only the crash type, and 10\% both—encouraging balanced reliance across modalities and enabling effective classifier-free guidance.

\subsection{Multi-Condition Classifier-Free Guidance}
\label{method_cfg}

Ctrl-Crash supports three conditioning modalities: initial image $c_I$, bounding box frames $c_B$, and crash type $c_T$. To enable independent control over each during inference, we extend classifier-free guidance (CFG) following \cite{liu2023compositionalvisualgenerationcomposable, pix2pix}, 
and denote the final noise estimate under multi-condition CFG as $\hat{\epsilon}_{\theta, \phi}$:

\vspace{-3mm}
\begin{equation}
\begingroup
\fontsize{8pt}{7.5pt}\selectfont
\begin{aligned}
  \hat{\epsilon}_{\theta, \phi}(\mathbf{x}_t, c_I, c_B, c_T) =\:
    &\epsilon_\phi(\mathbf{x}_t, c_I, \emptyset, \emptyset) \\
    &\!+ \gamma_B \left[ \epsilon_\theta(\mathbf{x}_t, c_I, c_B, \emptyset) 
            - \epsilon_\phi(\mathbf{x}_t, c_I, \emptyset, \emptyset) \right] \\
    &\!+ \gamma_T \left[ \epsilon_\theta(\mathbf{x}_t, c_I, c_B, c_T) 
            - \epsilon_\theta(\mathbf{x}_t, c_I, c_B, \emptyset) \right]
    \label{eqn:cfg_factor}
\end{aligned}
\endgroup
\end{equation}

Unlike standard CFG (Equation~\ref{eq:cfg_eqn}), which uses the same model for both conditional and unconditional predictions, we follow~\cite{uncond_priors} and decouple these roles: $\epsilon_\phi$ (the off-the-shelf SVD model) predicts the unconditional term, while $\epsilon_\theta$ (the fine-tuned Ctrl-Crash model) handles conditional terms. This avoids degraded unconditional priors and improves conditional fidelity. The coefficients $\gamma_B$ and $\gamma_T$ modulate the strength of bounding box and crash type conditionings, respectively, enabling flexible and interpretable control at inference.

\section{Experiments} 
\label{sec:results}
\subsection{Quantitative Evaluation}
\label{sec:quant_eval}

\begin{table}[!htb] \label{tab:1}
  \centering
  \vspace{-3mm}
  \caption{
    \textbf{Comparison of accident video generation quality across diffusion-based methods.} 
    We report FVD and JEDi scores (\(\downarrow\) lower is better). Scores marked with * are taken directly from the original papers and may not be strictly comparable due to differences in evaluation setup. The "Conditions" column describes inputs used as control signals. "img": initial image frame, "BB": bounding box frames, "text": textual description, "action": discrete crash category value. See Figure~\ref{fig:ctrlcrash_compare} for visuals.} 
    \vspace{1mm}
  \begin{tabular}{llcc}
    \toprule
    \textbf{Method} 
    & \textbf{Conditions} 
    & \textbf{FVD\(\downarrow\)}
    & \textbf{JEDi\(\downarrow\)} \\
    \midrule
    OAVD     & img + BB + text    &  5238* & - \\
    DrivingGen & img + BB + text  &  978.0* & - \\
    \midrule
    SVD base & img                  &  1420 & 3.628 \\
    AVD2     & img + text           &  1321 & 2.029 \\
    Ctrl-V   & img + BB     &  517.1 & 0.2910\\
    Ctrl-Crash  & img + BB + action  &  \textbf{449.5} & \textbf{0.1219} \\
    \bottomrule
  \end{tabular}
  \label{tab:generation_quality}
\end{table}

We evaluate the generation quality of Ctrl-Crash across two core settings: (1) general crash video generation in Table~\ref{tab:generation_quality}; and (2) varying the number of bounding box guidance frames provided in the prediction task to assess the impact of partial trajectory information in Table~\ref{tab:pred_qual}. All generated videos are 25 frames long at a resolution of $512 \times 320$, and metrics are computed over 200 samples unless otherwise specified.

We evaluate generation quality using several video and frame-level metrics. Fréchet Video Distance (FVD)~\cite{unterthiner2019fvd}  measures the similarity between the distributions of generated and real videos by analyzing video embeddings in I3D space under the assumption of Gaussian distributions; lower FVD values indicate a closer resemblance to real data. JEDi ~\cite{luo2025jedi} is a novel metric designed as an alternative to FVD by addressing its limitations through relaxing the assumption of Gaussian-distributed video feature embeddings, modifying the embedding space, and enabling convergence with a smaller sample size. In addition to these distributional-based video metrics, we report LPIPS (Learned Perceptual Image Patch Similarity)~\cite{zhang2018lpips}, SSIM (Structural Similarity Index), and PSNR (Peak Signal-to-Noise Ratio) in Table~\ref{tab:pred_qual}, which evaluate frame-level fidelity and perceptual closeness relative to ground truth, averaged over the video frames. These metrics provide complementary insights into how well the generated videos preserve appearance, structure, and temporal dynamics.

As shown in Table~\ref{tab:generation_quality}, Ctrl-Crash achieves the best results across both FVD and JEDi among compared methods, indicating strong alignment with real crash dynamics and superior video quality. For fair comparison, we compute the ground truth (GT) distribution for FVD and JEDi from a set of 500 randomly sampled MM-AU~\citep{fang2024abductive} validation videos, to accommodate for methods without GT alignment (e.g., AVD2) (see Appendix~F in supplementary material for more details). The SVD Base model, which serves as the foundation for Ctrl-Crash, performs poorly, as it was not trained for driving or crash-related content. Ctrl-V, while similar in architecture, lacks crash-specific training data and semantic control, leading to notable quality degradation near the crash event. AVD2 performs moderately well but exhibits inconsistent visual quality and weaker temporal coherence compared to Ctrl-Crash, as confirmed by FVD and JEDi. These results highlight the importance of both targeted training and structured control for crash simulation. 

\begin{table}[!htb]
  \centering
  \vspace{-3mm}
  \setlength{\tabcolsep}{5pt}
  \caption{\textbf{Effect of bounding box conditioning on crash video prediction quality.} 
    We evaluate Ctrl-Crash on the \textit{Crash Prediction} task by varying the number of initial bounding box frames provided as input (out of 25 total frames).}
    \vspace{1mm}
  \begin{tabular}{ccccccc}
    \toprule
    \textbf{\#BBs}
    & \textbf{FVD\(\downarrow\)}
    & \textbf{JEDi\(\downarrow\)}
    & \textbf{LPIPS\(\downarrow\)}
    & \textbf{SSIM\(\uparrow\)} 
    & \textbf{PSNR\(\uparrow\)} \\
    \midrule
    0   &  422.1 & 0.3155 & 0.3856 & 0.5188 & 16.57 \\
    3   &  375.7 & 0.2949 & 0.3594 & 0.5434 & 17.27 \\
    9   &  353.3 & 0.2160 & 0.3392 & 0.5614 & 17.83 \\
    25  &  \textbf{323.9} & \textbf{0.1219} & \textbf{0.3113} & \textbf{0.5836} & \textbf{18.33} \\
    \bottomrule
  \end{tabular}
  \label{tab:pred_qual}
\end{table}

We also study the impact of varying the number of bounding box frames used as conditioning for Ctrl-Crash, in Table~\ref{tab:pred_qual}. Specifically, we compare using partial bounding-box frame conditioning from zero frames to a few bounding box frames (\textit{Crash Prediction} task) all the way to a fully defined motion sequence by providing all 25 frames (\textit{Crash Reconstruction} task). As shown in Table~\ref{tab:pred_qual}, generation quality improves consistently with the number of provided bounding box frames. This trend is visible across both distributional metrics (FVD and JEDi) and frame-level scores (LPIPS, SSIM, PSNR), and supports the hypothesis that denser spatial constraints lead to easier prediction tasks and more stable outputs. The results validate that Ctrl-Crash gracefully interpolates between unconditional prediction and fully supervised reconstruction. 

Across all benchmarks, Ctrl-Crash delivers significant improvements over prior diffusion-based crash generation models. It handles both unconstrained and highly conditioned inputs, demonstrating its utility for generating diverse crash outcomes and precise reconstructions. In the next section, we complement these quantitative results with a user study and qualitative visualizations.

\subsection{Qualitative Evaluation}

\begin{figure}[!htb]
  \centering
  \vspace{-3mm}
  \includegraphics[width=\linewidth]{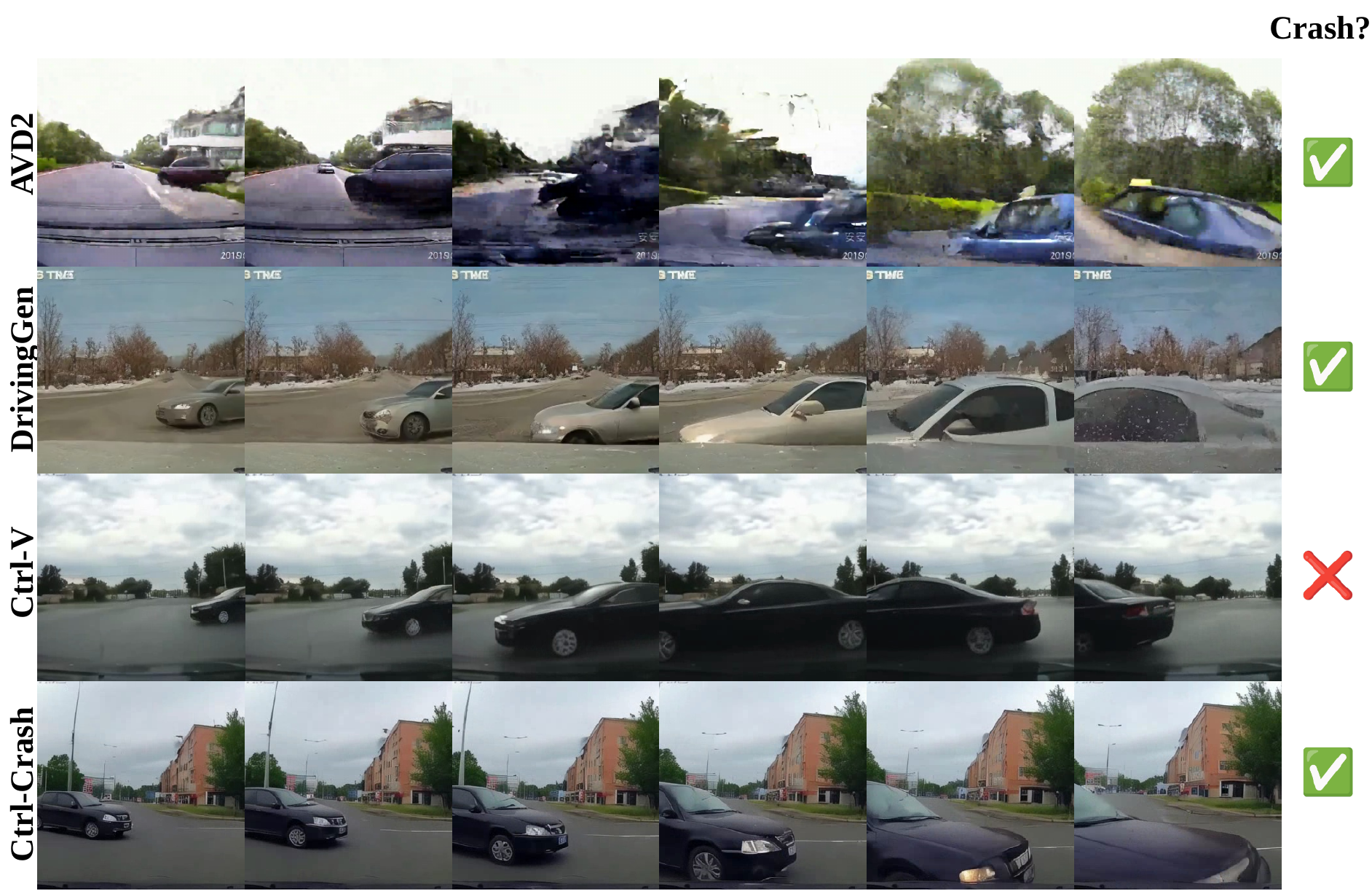}
  \caption{\textbf{Qualitative comparison of AVD2, DrivingGen, Ctrl-V, and Ctrl-Crash}. Ctrl-Crash achieves superior visual fidelity and realistic crash dynamics, outperforming baselines that suffer from low quality, inconsistency, or lack of plausible crash events.}
  \label{fig:ctrlcrash_compare}
\end{figure}

\begin{figure}[!htb]
  \includegraphics[width=\linewidth]{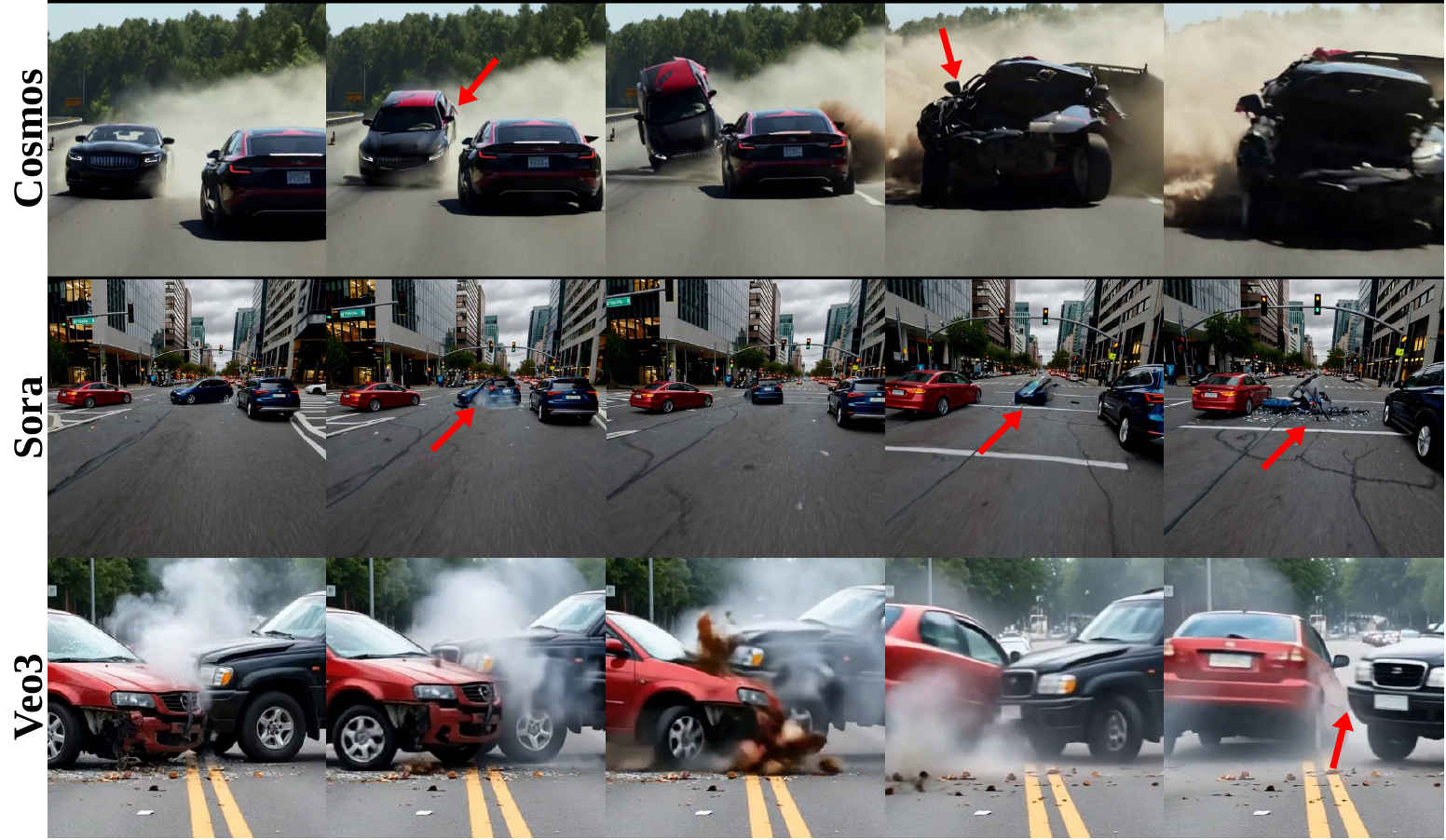}
  \caption{\textbf{State-of-the-art text-to-video diffusion models struggle to generate a realistic car accident}. \textit{Top}: Nvidia Cosmos (7B Text2World),  \textit{Middle}: OpenAI Sora, \textit{Bottom}: DeepMind Veo3.}
  \label{fig:others_fail_main}
\end{figure}

We qualitatively tested leading video diffusion models—OpenAI's Sora~\cite{brooks2024videoworldsimulators_sora}, Nvidia's Cosmos~\cite{agarwal2025cosmos}, and DeepMind's Veo3~\cite{deepmind_veo3}—for their ability to generate car crash scenarios. While these models produce high-resolution, cinematic visuals, they consistently struggle to depict physically plausible accidents. In our experiments, Cosmos and Sora failed to generate convincing crashes across several prompt variations, while Veo3, which we tested with a single sample, showed limited realism despite momentarily resembling a crash (see \ref{appendix:more_qual} for exact prompts). These examples, shown in Figure~\ref{fig:others_fail_main}, suggest that current general-purpose models lack the exposure, the fine-grained control, and physical grounding required for realistic accident simulation.

Additionally, we conducted a brief user study to evaluate visual quality and physical realism of the generated videos based on human preference. Results show clear and significant preference for Ctrl-Crash among similar methods. Results and details are presented in \ref{appendix:survey}.

\section{Conclusion}
In this work, we introduced Ctrl-Crash, a controllable video diffusion framework that generates realistic car crash scenarios from a single frame, which achieves state-of-the-art performance among diffusion-based methods designed for crashes, and enables counterfactual reasoning by varying spatial and semantic control inputs. To support training and evaluation, we also developed a processing pipeline for extracting bounding boxes from crash videos and released curated, annotated versions of MM-AU, RussiaCrash, and BDD100k to facilitate future research in crash simulation and generative modeling.

Despite its strong performance, our approach has several limitations, motivating promising directions for future work. Counterfactual outcomes can be difficult to generate when the initial scene strongly suggests a different crash outcome than the one being conditioned on. The model also relies heavily on 2D bounding boxes, which makes it sensitive to tracking errors—particularly in fully conditioned reconstruction. Moreover, 2D boxes often lack information about agent orientation and rotation, limiting realism for behaviors such as spinouts or rollovers. Future extensions may explore 3D bounding boxes or richer trajectory representations to overcome this. Another limitation is the use of a fixed set of five discrete crash type labels to define semantic intent. While simple and effective, this formulation constrains expressiveness. Incorporating natural language as a conditioning signal could unlock more nuanced control and allow the model to interpret detailed scene descriptions or intentions that go beyond predefined categories and bounding-box information. We envision Ctrl-Crash as a foundation for building more general, interpretable, and controllable generative models for safety-critical autonomous driving research.
\vspace{-3mm}\\

\noindent\textbf{Acknowledgments:}
We thank Samsung, the IVADO and the Canada First Research Excellence Fund (CFREF) / Apogée
Funds, the Canada CIFAR AI Chairs Program, and the NSERC Discovery Grants program for
financial support. We also thank Mila - the Quebec AI Institute for compute resources.
\vspace{-5mm}





\newgeometry{left=2.5cm, right=2.5cm, top=2.5cm, bottom=2.5cm} 
\onecolumn
\appendix

\section{Additional Results}

\subsection{Additional Quantitative Results: Ablations}
\label{appendix:more_quant}

We present in Table~\ref{tab:generation_quality_ablations} additional results to complement results given in Table~1 of the main paper. In addition to FVD~\cite{unterthiner2019fvd} and JEDi~\cite{luo2025jedi} values, we present LPIPS~\cite{zhang2018lpips}, SSIM, and PSNR for Ctrl-Crash and its ablations. Ctrl-V is similar to Ctrl-Crash but has not been trained on crash data and was not designed with dynamic crash video generation in mind. SVD base is the base model Ctrl-Crash is derived from through finetuning. For Ctrl-Crash and Ctrl-V the full sequence of bounding boxes is given, so we are expecting a reconstruction similar to the ground-truth crash. Across all metrics, we see that Ctrl-Crash performs the best at reconstructing plausible crashes.

\begin{table}[h!]
  \caption{Generated accident video quality compared to baseline ablations methods}
  \label{sample-table}
  \centering
  \begin{tabular}{llccccc}
    \toprule
    \textbf{Method}     
    & \textbf{Conditions}   
    & \textbf{FVD\(\downarrow\)}      
    & \textbf{JEDi\(\downarrow\)} 
    & \textbf{LPIPS\(\downarrow\)} 
    & \textbf{SSIM\(\uparrow\)} 
    & \textbf{PSNR\(\uparrow\)} \\
    \midrule
    SVD base & img                  & 1420  & 3.628 & 0.5800 & 0.3074 & 11.74 \\
    Ctrl-V   & img + bbox           & 517.1 & 0.2910 & 0.3670 & 0.5372 & 16.81 \\
    Ctrl-Crash (Ours)     & img + bbox + action  & \textbf{449.5} & \textbf{0.1219} & \textbf{0.3113} & \textbf{0.5836} & \textbf{18.33} \\
    \bottomrule
  \end{tabular}
  \label{tab:generation_quality_ablations}
\end{table}

\begin{table}[!htb]
  \centering
  \vspace{-3mm}
  \caption{\textbf{Effect of crash type conditioning on crash video generation quality.} We evaluate Ctrl-Crash on the \textit{Crash Counterfactuals} task by varying the crash type conditioning and nothing else. For each case, 200 videos were generated using the same initial image and three bounding box frames as initial context, but with different desired crash types.} 
  \vspace{1mm}
  \begin{tabular}{llcc}
    \toprule
    \textbf{Crash Type}
    & \textbf{FVD\(\downarrow\)}
    & \textbf{JEDi\(\downarrow\)} \\
    \midrule
    GT crash type & 375.7 & 0.2949 \\
    \midrule
    0 - no crash &  400.9 & 0.4514 \\
    1 - ego-only &  379.5 & 0.3091 \\
    2 - ego/vehicle &  372.9 & 0.3001\\
    3 - vehicle-only &  398.1 & 0.3856 \\
    4 - vehicle/vehicle &  383.4 & 0.3416 \\
    \bottomrule
  \end{tabular}
  \label{tab:counter_qual}
\end{table}

In Table~\ref{tab:counter_qual}, we evaluate Ctrl-Crash on the \textit{Crash Counterfactuals} task by varying only the crash type while keeping the initial image and partial bounding box sequence fixed. The model achieves its best performance on ego/vehicle crashes (type 2), likely due to their higher frequency in the training data, while vehicle-only crashes (type 3) are harder to generate plausibly, reflecting their visual complexity. Interestingly, ego-only crashes (type 1), though underrepresented in the dataset, also perform well—possibly because they involve simpler dynamics without visible impact. Overall, counterfactual generations remain close in quality to the ground-truth-guided baseline, suggesting that Ctrl-Crash can produce realistic alternate outcomes, though performance declines when strong visual cues in the initial frame conflict with the intended crash type. Class frequency in the training set are provided in Appendix~\ref{appendix:datasets}.

\subsection{Additional Qualitative Results}
\label{appendix:more_qual}

\paragraph{SOTA models}

\begin{figure}[!htb]
  \includegraphics[width=\linewidth]{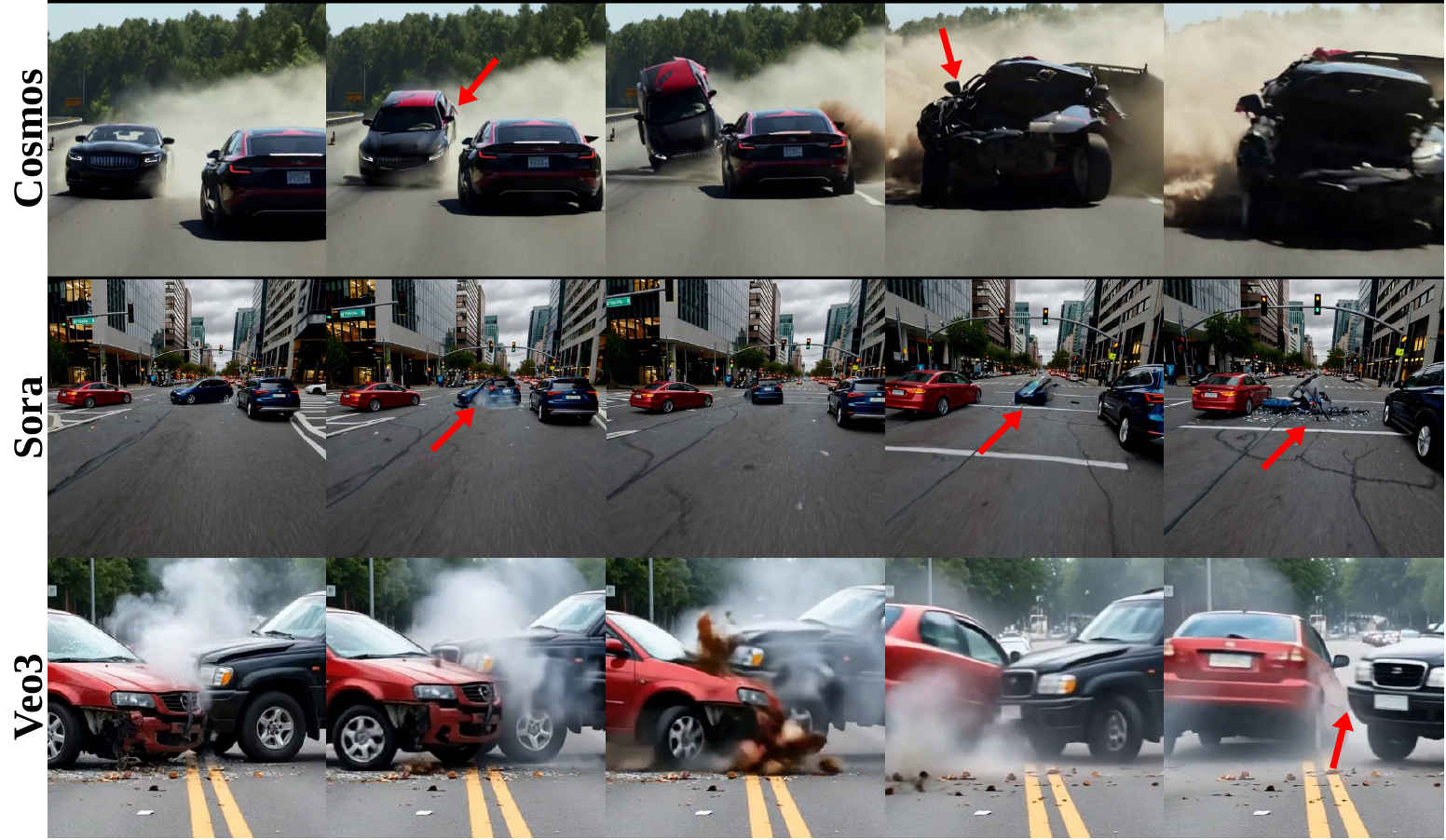}
  \caption{\textbf{State-of-the-art text-to-video diffusion models struggle to generate a realistic car accident} (as shown in main paper). \textit{Top}: Nvidia Cosmos (7B Text2World),  \textit{Middle}: OpenAI Sora, \textit{Bottom}: DeepMind Veo3.}
  \label{fig:others_fail}
\end{figure}

State-of-the-art diffusion-based video models, such as OpenAI's Sora~\cite{brooks2024videoworldsimulators_sora}, Nvidia's Cosmos~\cite{agarwal2025cosmos}, and  DeepMind's Veo3~\cite{deepmind_veo3} fail to generate plausible crash scenarios. We tested these three models with several samples and though they tend to generate high definition and smooth video, they consistently fail to generate physically plausible crashes (if any crash at all), we show some examples in Figure~\ref{fig:others_fail}.

In the top row of the figure we show a sample result from Nvidia Cosmos-Predict1-7B-Text2World that was instructed with the following text prompt: \textit{"On a highway two cars collide at very fast speeds head-on"}. This produces a highly implausible scene, where the car on the left suddenly starts to levitate from its rear-end and a cloud of smoke resembling an explosion appears, followed by disjointed fragments of torn metal emerging from the ground that transforms in a dark vehicle rushing towards the camera. 

The Nvidia Cosmos model suite has been specifically trained for physics-aware generation, but still fail to generate a car crash. Alternatively, we show a sample from OpenAI's Sora video model in Figure~\ref{fig:others_fail} in the middle row. The Sora video model was given the prompt: \textit{"At an intersection two cars collide with each other at full speed resulting in a crash"}. The resulting video shows a car that spins erratically, changes shape and direction, and produces visible artifacts—culminating in a pile of twisted metal and glass. 

Finally, we show an example using DeepMind's Veo3 in the bottom row of Figure~\ref{fig:others_fail}. We used the following prompt: \textit{"At an intersection two cars collide with each other at full speed resulting in a crash"}. Perhaps the closest to depicting a car crash when looking at individual frames, but when watching the video as a whole it becomes clear that the events are completely unrealistic. In the generated video, we see two still vehicles very close to each other that start accelerating, then we see debris and smoke as the cars slowly move towards each other, then the front of the cars phase through each other and reveal completely undamaged cars, and finally the red car drives away from the scene.

In contrast, through our approach of sourcing crash data, preprocessing and annotating sequences with bounding boxes, crash type conditioning, and our stochastic conditioning approach, Ctrl-Crash can much more reliably generate physically plausible crash imagery and collision dynamics.

\begin{figure}[!htb]
  \includegraphics[width=\linewidth]{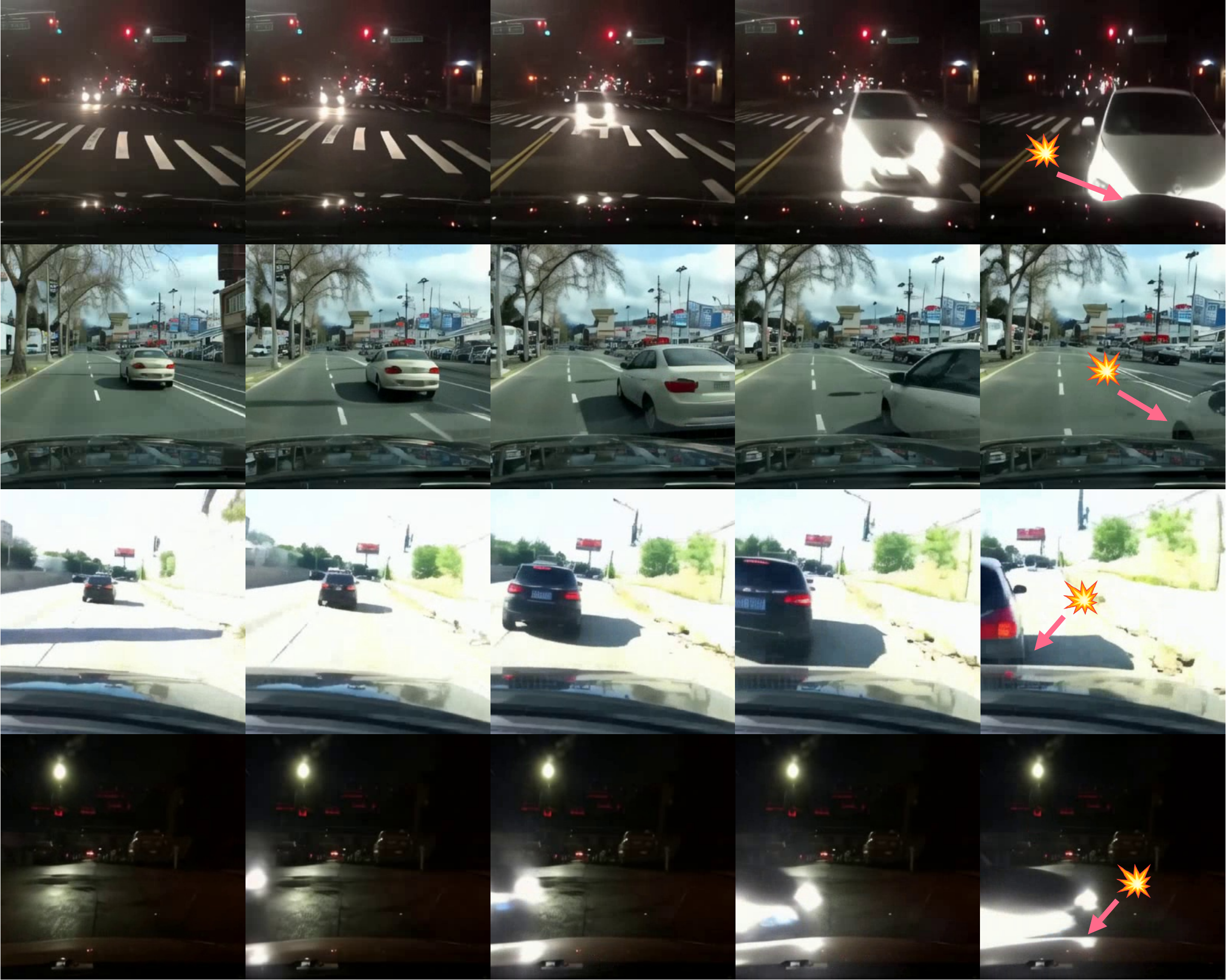}
  \caption{Crash scenarios generated by Ctrl-Crash from non-crash scenes in the BDD100k dataset. Each row shows 5 frames from a generated 25-frame clip. Despite originating from benign driving scenes, the model produces visually coherent and diverse crash outcomes across examples.}
  \label{fig:bdd_examples}
\end{figure}

\paragraph{Generating Crashes from Non-Crash Data} To demonstrate the practical utility of Ctrl-Crash for safety-critical dataset augmentation, we apply our model to generate crash scenarios seeded from the BDD100k dataset, an established real-world driving dataset that contains no actual crashes. By conditioning on initial frames and coarse agent trajectories from BDD100k, Ctrl-Crash is able to hallucinate plausible accident outcomes, transforming otherwise uneventful driving clips into diverse crash scenarios. This capability is especially valuable for enriching datasets used in autonomous driving research, where real crash data is scarce, sensitive, or ethically challenging to collect. Synthesizing rare and hazardous events from benign scenes enables safer training and evaluation of perception and planning systems without requiring exposure to real-world danger. In Figure~\ref{fig:bdd_examples}, we show some examples of car crash videos generated by Ctrl-Crash from non-crash scenes in the BDD100k dataset.

\begin{figure}[htbp]
  \centering
  \makebox[\textwidth][c]{\includegraphics[width=\linewidth]{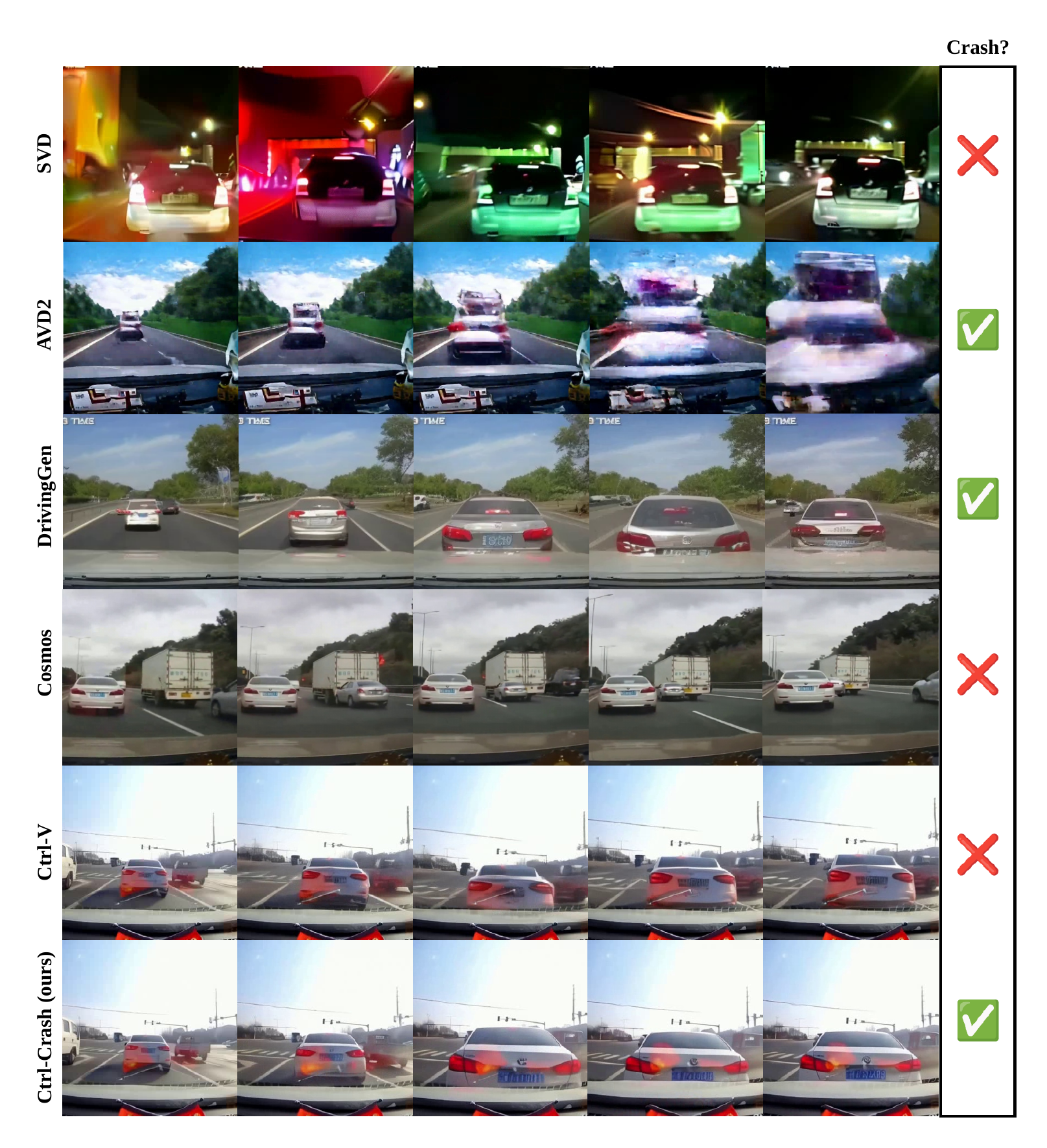}}
  \caption{Qualitative comparison of "rear-end crashes" between different methods. For each method, we show 5 frames from the video along with either a green check mark if there appears to have a crash in the video otherwise a red 'X'. \textbf{From top to bottom}: \textbf{SVD} (stable-video-diffusion-img2vid)~\cite{blattmann2023svd} prompted with the initial frame from a rear-end crash video, we see some normal driving but very inconsistent lighting and color shades with visible distorsions. \textbf{AVD2}~\cite{li2025avd2}, we see what appears to be a rear-end crash with a very distorted leading vehicle and background. \textbf{DrivingGen}~\cite{guo2024drivinggen}, we see a rear-end crash with a leading vehicle that changes appearance every frame. Overall the video is very choppy with little temporal consistency. \textbf{Cosmos} (Cosmos-Predict1-7B-Video2World)~\cite{agarwal2025cosmos}, prompted with text suggesting a rear-end crash and 9 initial images where a car is rapidly approaching a truck, the predicted frames show the car unrealistically shrinking as it approaches the truck without any signs of a collision. \textbf{Ctrl-V}~\cite{luo2025ctrlv}, prompted with a sequence of bounding-boxes suggesting a rear-end crash with a leading car, we see the leading car keep its distance and not crash occurs. \textbf{Ctrl-Crash} (ours): prompted with the same bounding box sequence as Ctrl-V and with the discrete crash type "ego/vehicle crash", we see a physically plausible rear-end collision with the ego vehicle visibly shaking from the impact.}
  \label{fig:qual_rear_end}
\end{figure}

\begin{figure}[htbp]
  \centering
  \makebox[\textwidth][c]{\includegraphics[width=\linewidth]{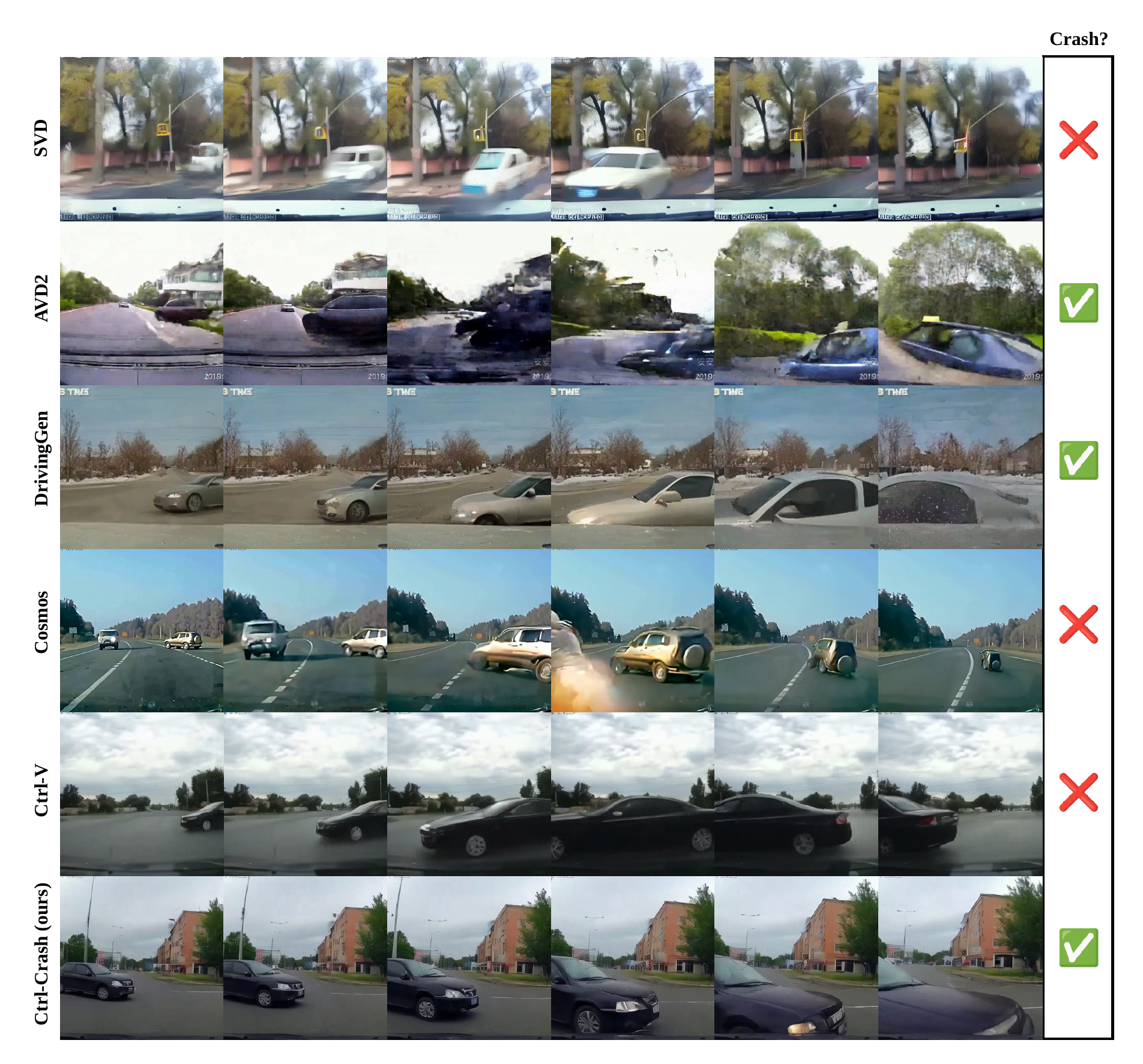}}
  \caption{Qualitative comparison of "t-bone crashes" between different methods. For each method, we show 6 frames from the video along with either a green check mark if there appears to have a crash in the video otherwise a red 'X'. \textbf{From top to bottom}: \textbf{SVD} (stable-video-diffusion-img2vid)~\cite{blattmann2023svd} prompted with the initial frame from a t-bone car crash video, we see blurry vehicle a distorted motion blur as it drives in front of the ego vehicle without any collision. \textbf{AVD2}~\cite{li2025avd2}, we can make out what seems to be a t-bone crash with a heavily distorted black car. There are many artifacts and temporal inconsistencies which makes the sequence of events hard to follow. \textbf{DrivingGen}~\cite{guo2024drivinggen}, a gray sedan drive in front of the ego vehicle progressively getting closer until it seems to collide with it. Motion is jerky and uneven between timesteps and the appearance of the gray car shapes almost every frame. \textbf{Cosmos} (Cosmos-Predict1-5B-Video2World)~\cite{agarwal2025cosmos}, prompted with creating a t-bone crash and 9 initial frames showing a car turn in front of the ego car, we see the leading car start to distort as the ego approaches it and then it shrivels and shrinks until it nearly disappears. \textbf{Ctrl-V}~\cite{luo2025ctrlv}, prompted with a sequence of bounding-boxes suggesting a t-bone crash with a car incoming from the left, we see a car drive in from the left and then just passed the ego car without any collision. \textbf{Ctrl-Crash} (ours): prompted with the same bounding box sequence as Ctrl-V and with the discrete crash type "ego/vehicle crash", we see a physically plausible t-bone collision with the a black sedan incoming from the left.}
  \label{fig:qual_tbone}
\end{figure}

\begin{figure}[htbp]
  \centering
  \makebox[\textwidth][c]{\includegraphics[width=\linewidth]{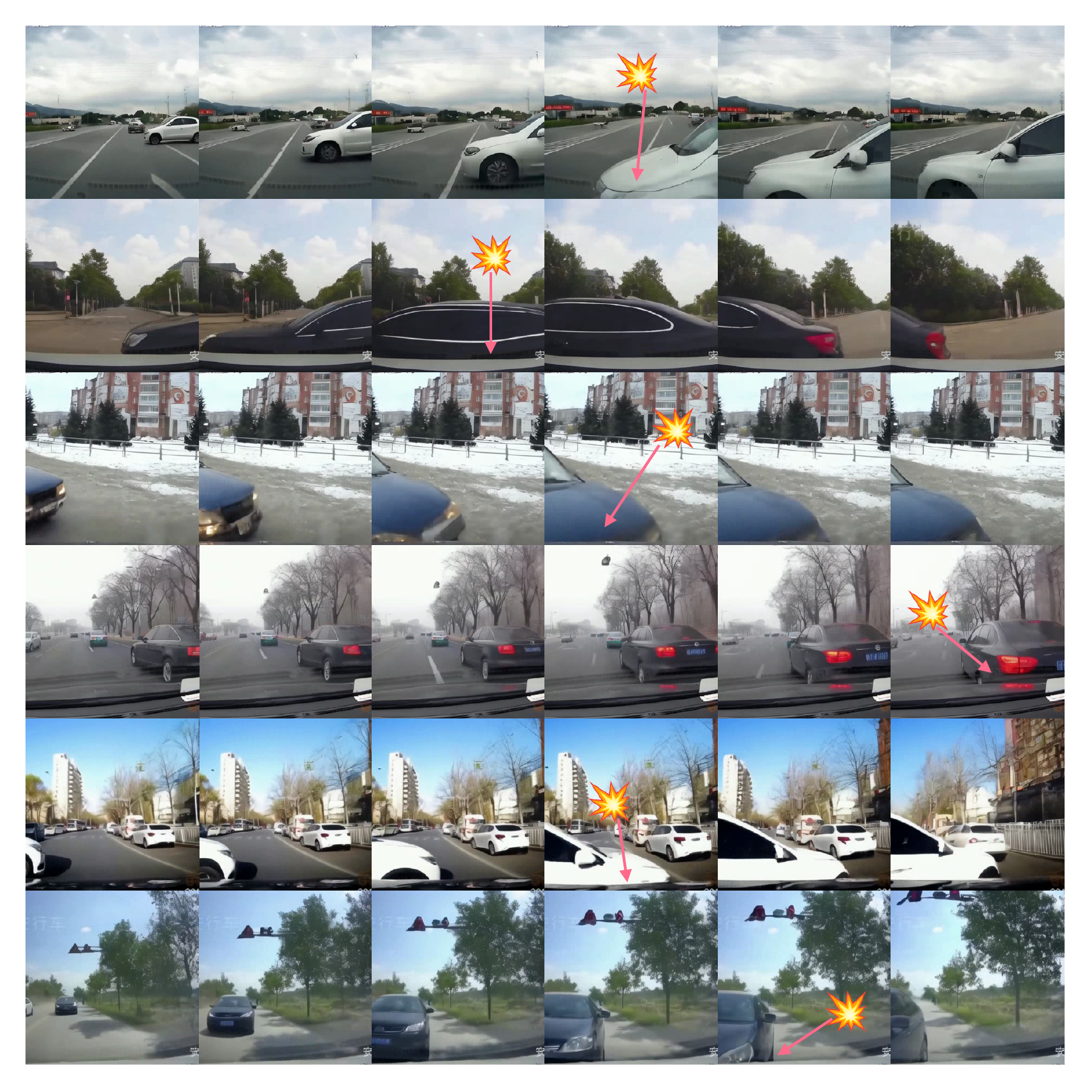}}
  \caption{Ctrl-Crash qualitative results conditioned on an initial 9 bounding box frames (i.e., the first two frames of these sequences were conditioned on bounding box frames, but not the others).}
  \label{fig:qual_9bboxes}
\end{figure}

\begin{figure}[htbp]
  \centering
  \makebox[\textwidth][c]{\includegraphics[width=\linewidth]{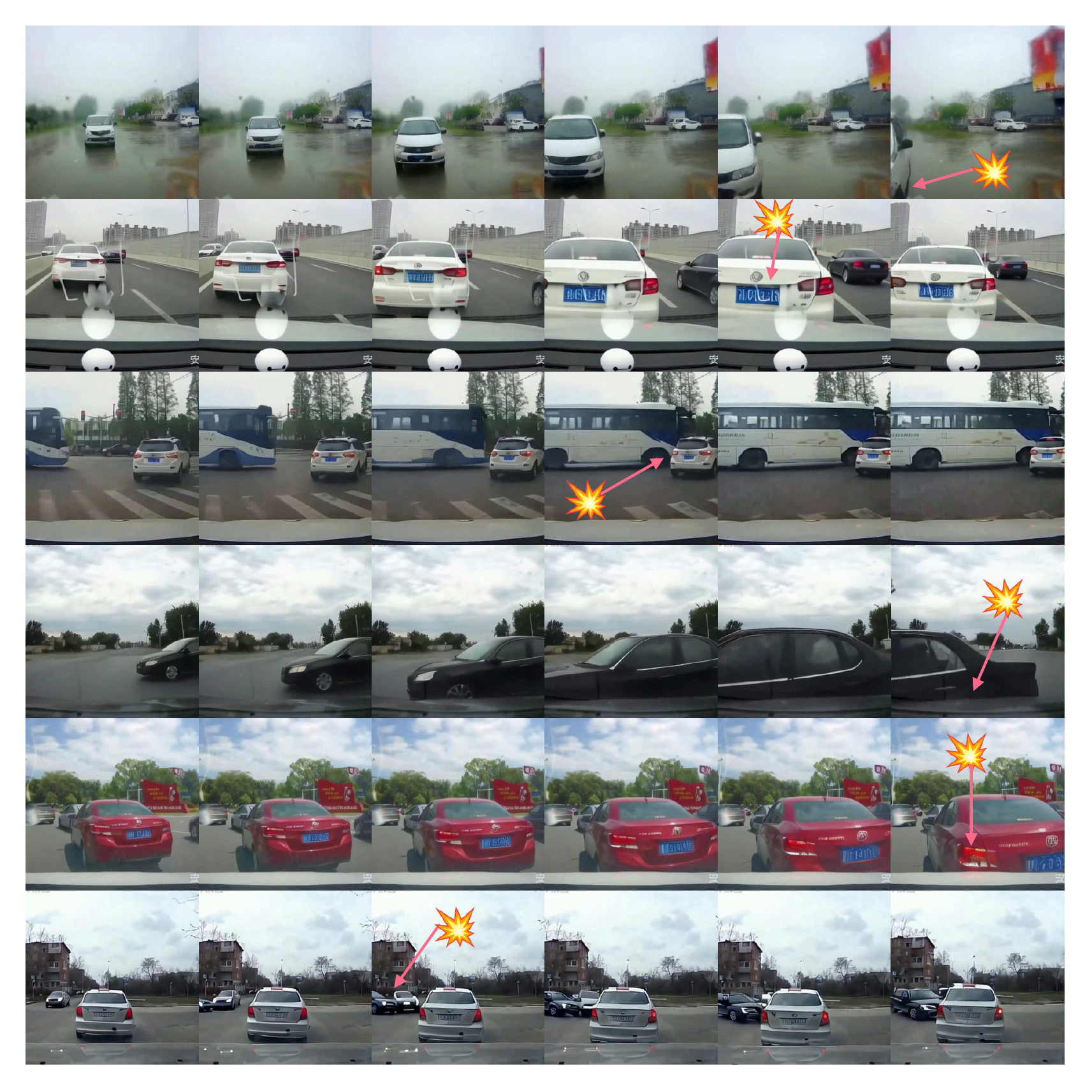}}
  \caption{Ctrl-Crash qualitative results conditioned on 25 (all) bounding box frames.}
  \label{fig:qual_allbboxes}
\end{figure}

\paragraph{Car Crash Video Diffusion Models Comparison}
Other recent work \cite{guo2024drivinggen, li2025avd2} focus on the task of car crash video generation, but lack the visual quality for convincing car crashes. As mentioned in the main paper (and presented in Appendix~\ref{appendix:survey}), we conducted a user survey that shows that Ctrl-Crash significantly outperforms these existing methods for visual fidelity and physical plausibility of generated car crashes. We show additional qualitative comparisons in Figure~\ref{fig:qual_rear_end} and Figure~\ref{fig:qual_tbone}, along with many samples generated from Ctrl-Crash in Figure~\ref{fig:qual_9bboxes}, and Figure~\ref{fig:qual_allbboxes}.

\clearpage
\section{Low Resolution Filtering Heuristic}
\label{appendix:fft}

\begin{algorithm}[htbp]
\caption{Estimate Upscaling Factor from Image}
\label{alg:upsizing_factor}
\begin{algorithmic}[1]
\Procedure{EstimateUpsizingFactor}{\texttt{image\_path}}
    \State Load image in grayscale: $I \gets \texttt{cv2.imread(image\_path, GRAYSCALE)}$
    \State Compute 2D FFT and shift: $F \gets \texttt{np.fft.fftshift(np.fft.fft2}(I))$
    \State Compute magnitude spectrum: $M \gets |\!|F|\!|$
    \State Get image size: $(h, w) \gets \texttt{shape}(I)$
    \State Center: $(c_x, c_y) \gets (w // 2, h // 2)$
    \State Define low-frequency radius: $r \gets \min(c_x, c_y) // 4$
    \State Create circular mask $L$ of radius $r$ centered at $(c_x, c_y)$
    \State Compute high-frequency mask: $H \gets 1 - L$
    \State Compute high-frequency energy: $E_{\text{high}} \gets \sum M \cdot H$
    \State Normalize energy: $e \gets E_{\text{high}} / (h \cdot w)$
    \State Compute upscaling factor: $f \gets 1 / (1 + e)$
    \State \Return $f$
\EndProcedure
\end{algorithmic}
\end{algorithm}

A key challenge in curating video datasets from online sources is ensuring that the selected samples reflect genuine high-resolution content rather than upscaled of heavily compressed footage (via resizing). This is particularly relevant for dashcam videos, where visual clarity and high-frequency details are critical for learning accurate dynamics and semantics. As part of the video preprocessing pipeline (see Section~\ref{appendix:video_proc}), we introduce a frequency-based filtering heuristic to estimate the degree of upscaling in video frames.

The heuristic is based on the observation that upscaled or low-quality images tend to exhibit diminished high-frequency content due to interpolation artifacts and smoothing. By analyzing the energy distribution in the frequency domain, we define an \textit{upsizing factor} that acts as a proxy for the likelihood of artificial upscaling. The method proceeds as follows:

\begin{enumerate}
    \item \textbf{Image Loading.} Each video frame is converted to grayscale to simplify analysis and focus on structural image content rather than color channels.

    \item \textbf{Fourier Transform.} A 2D Fast Fourier Transform (FFT) is applied to the grayscale image to obtain its frequency representation. The spectrum is shifted so that low-frequency components are centered.

    \item \textbf{Magnitude Spectrum.} The magnitude of the frequency components is computed by taking the absolute value of the FFT coefficients. This magnitude spectrum represents the intensity of different spatial frequencies in the image.

    \item \textbf{High-Frequency Energy Calculation.} A circular mask is applied to exclude the low-frequency region in the center of the spectrum. The sum of magnitudes outside this central region defines the high-frequency energy.

    \item \textbf{Normalization.} The high-frequency energy is normalized by the total number of pixels to produce a scale-invariant energy score.

    \item \textbf{Upsizing Factor Estimation.} The final upsizing factor $U$ is computed as:
    $U = \frac{1}{1 + E}$,
    where $E$ is the normalized high-frequency energy. This formulation ensures that frames with low high-frequency energy—indicative of potential upscaling—yield lower scores.

    \item \textbf{Interpretation.} The resulting upsizing factor provides a heuristic score: lower values indicate a higher likelihood of artificial enlargement or blur, while higher values suggest more genuine, detailed high-resolution frames.

\end{enumerate}

This frequency-domain metric serves as a lightweight yet effective automated quality check, helping to filter out training samples that lack meaningful visual detail. The corresponding implementation is summarized formally in Algorithm~\ref{alg:upsizing_factor}.

\clearpage
\section{Video Processing}
\label{appendix:video_proc}
\paragraph{Video Processing} To train our model, we use the MM-AU dataset~\cite{fang2024abductive}, a large-scale collection of in-the-wild dashcam crash videos sourced from public platforms. However, due to the variable quality and content of these videos, we curate the dataset using a multi-stage filtering and preprocessing pipeline. The process can be summarized as follows:
\begin{enumerate}
    \item Filtering out low-quality, compressed, or blocky videos based on the low-resolution heuristics (described in Appendix~\ref{appendix:fft}).
    \item Removing clips with shot changes or unnatural scene cuts using scene detection.
    \item Normalizing the format: frame rate, resolution, clip length.
    \item Excluding scenes involving visible humans (e.g., pedestrians, cyclists, motorbikes).
\end{enumerate}

In the first step, we apply frequency-domain heuristics to identify and remove videos that suffer from excessive compression artifacts or poor resolution. These issues can hinder the model’s ability to learn coherent motion and object dynamics. We use a Fast Fourier Transform (FFT)-based method to detect videos with large blocky regions or low-frequency dominance—signals of strong compression or blur. This approach, described in Appendix~\ref{appendix:fft}, helps prioritize clips with high visual clarity and well-defined agent motion.

Next, we use PySceneDetect~\cite{PySceneDetect} to identify and discard videos containing abrupt scene transitions or camera disruptions—such as sudden viewpoint changes or dashcams that fall mid-recording. Removing such discontinuities ensures more stable temporal consistency within each training clip.

We then normalize all video clips to a consistent resolution of $512\times 320$ (width × height) and sample them at 6 frames per second. This not only aligns with the spatial and temporal input requirements of our model but also helps crop out unwanted overlays such as watermarks that often appear near video borders. We segment each video into fixed-length clips of 25 frames (approximately 4 seconds), discarding any segments shorter than this threshold.

Finally, for ethical and safety considerations, we exclude all videos depicting visible humans involved in crashes. This includes scenes involving pedestrians, cyclists, or motorcyclists. Our intention is to avoid training the model to depict harmful or distressing scenarios involving vulnerable road users.

In addition to the automated filtering steps, we also perform manual review of many video samples using internal tooling (made available on the project GitHub). This step helps identify edge cases and poor-quality examples that may have bypassed the automated pipeline, ensuring a cleaner and more appropriate training set.

\clearpage
\section{YOLO-SAM: Hybrid Bounding Box Annotation Pipeline}
\label{appendix:bbox_gen}

\begin{figure}[h!]
  \centering
  \makebox[\textwidth][c]{\includegraphics[width=\linewidth]{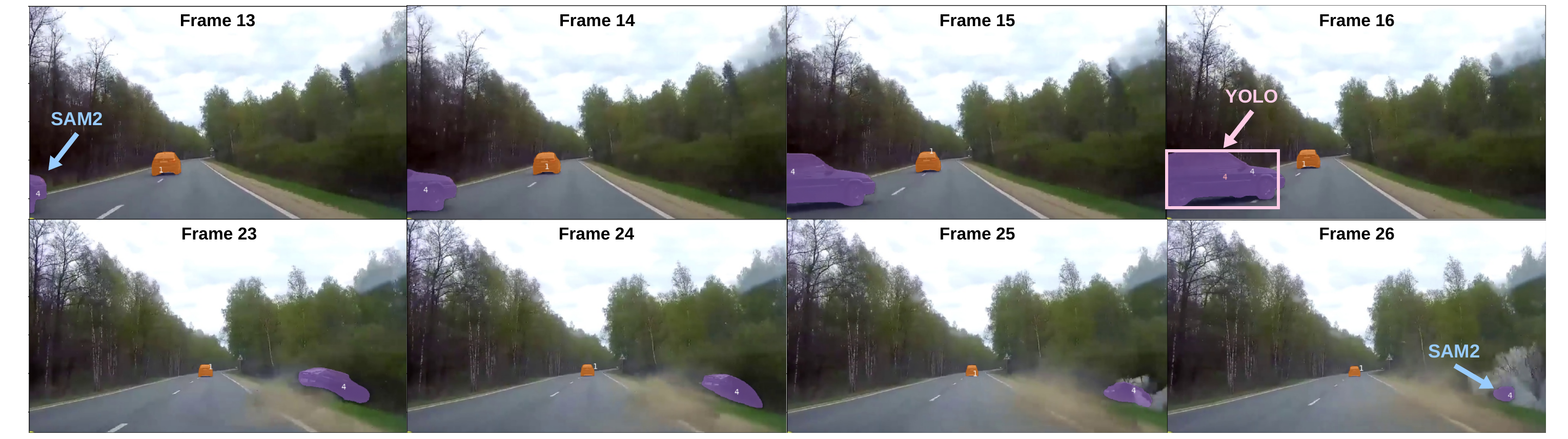}}
  \caption{Example of bounding box tracking for a vehicle veering off the road. SAM2 segmentations for the two vehicles are shown in purple and orange. YOLOv8 detects the purple car only starting at frame 16 (light pink box), while SAM2 successfully infers its presence across many more frames by propagating the segmentation bidirectionally in time. This provides more complete and temporally consistent annotations, especially in cases where objects become partially occluded or deformed.}
  \label{fig:sam_yolo_figure}
\end{figure}

We develop a hybrid annotation pipeline, YOLO-SAM, to generate high-quality bounding boxes for visible road users in crash and driving videos. This combines the speed and class-awareness of YOLOv8~\cite{yolo2023} with the instance-level segmentation and temporal propagation capabilities of SAM2~\cite{ravi2024sam2}. The hybrid design addresses detection failures due to occlusion, deformation, or sudden motion—common in crash scenarios—and supports dynamic entry/exit of agents.

YOLOv8 is run frame-by-frame to detect standard object classes (e.g., cars, trucks, buses) and assign initial bounding boxes and track IDs. While fast and accurate under normal conditions, YOLO has several shortcomings:
\begin{enumerate}
    \item \textbf{Duplicate Detections:} YOLO can assign multiple detections to the same object over time. We reject predictions if their IoU with previous detections exceeds 0.8.

    \item \textbf{Tracking Loss:} YOLO may lose an object (e.g., under occlusion) and redetect it with a new track ID. These failures are corrected using SAM2.
\end{enumerate}

To address YOLO’s limitations, we use SAM2 to refine and extend YOLO detections:
\begin{enumerate}
    \item \textbf{Shape Correction:} SAM2 provides temporally consistent masks, reducing YOLO’s tendency to distort box sizes and shapes.

    \item \textbf{Redetection Verification:} When YOLO redetects a lost object with a new ID, we compare its spatial overlap with SAM2’s mask to either restore the original ID or assign a new one.

    \item \textbf{Track ID Switches:} If YOLO mistakenly assigns an existing ID to a new object, SAM2's predictions are used to detect the mismatch and correct the ID.

    \item \textbf{Early-frame Recovery:} SAM2 supports bidirectional propagation, allowing recovery of early frames missed by YOLO. However, to avoid hallucinated presence, we only accept early-frame completions up to a few frames before YOLO’s first detection, and only when confidence is high.
\end{enumerate}

Figure~\ref{fig:sam_yolo_figure} shows an example where a vehicle exits the road and YOLO detects it only mid-sequence, while SAM2 successfully tracks it throughout by propagating the mask temporally.

For each YOLO-detected object, SAM2 is prompted to generate per-frame instance masks. These masks are converted to tight bounding boxes, and tracking continuity is enforced by ID consistency checks based on IoU overlap. The result is a per-frame annotation of bounding boxes with consistent track IDs and object classes.

Compared to YOLO or SAM2 alone, the hybrid approach provides significantly more complete and robust annotations, crucial for training controllable video generation models like Ctrl-Crash. The full implementation is available in our project repository for reproducibility.

\clearpage
\section{Bounding Box Conditioning}
\label{appendix:bbox_viz}

\begin{figure}[h!]
  \centering
  \makebox[\textwidth][c]{\includegraphics[width=\linewidth]{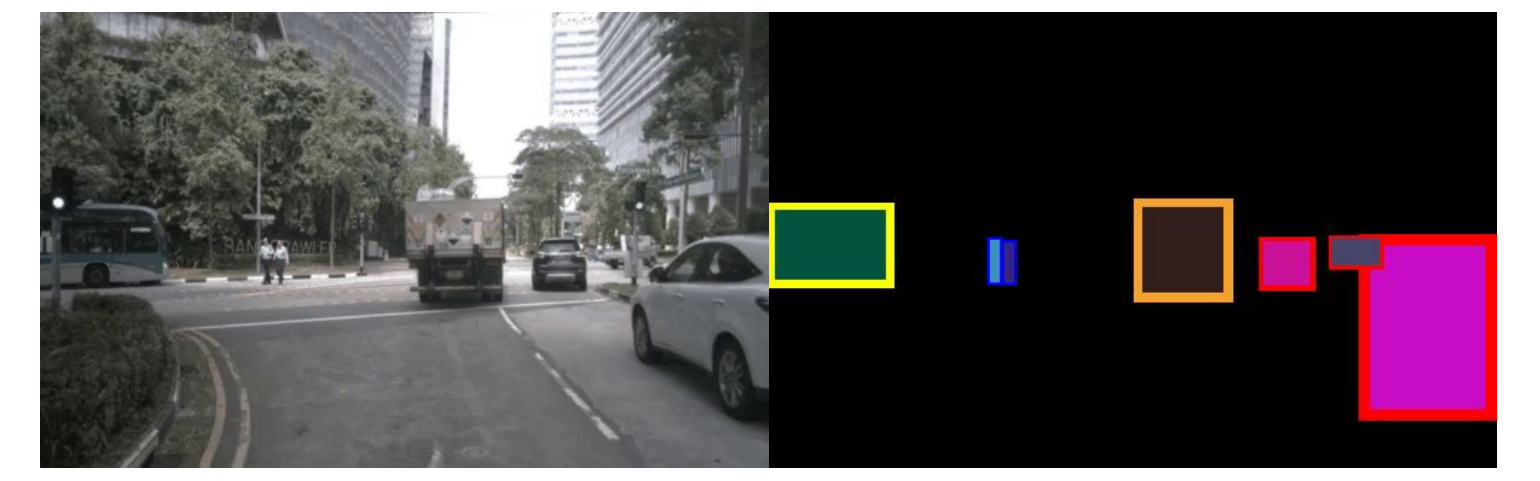}}
  \caption{\textbf{Left}: example frame from a driving video. \textbf{Right}: associated bounding box frame conditioning generated from our pipeline. Road users are represented as 2D bounding boxes with unique fill colors representing their track ID and specific border colors representing their class.}
  \label{fig:bbox_example}
\end{figure}

To provide spatial guidance to the generative model, we convert the bounding box trajectories of road users into RGB control frames that serve as conditioning input to the ControlNet. An example of such a control frame, alongside its corresponding real image, is shown in Figure~\ref{fig:bbox_example}. Each control frame encodes the complete set of bounding boxes for a single timestep using a color-coded rasterization scheme that encodes both object identity and semantic class.

\paragraph{Track ID Encoding (Fill Color)}
Each bounding box is filled with a unique RGB color derived from its object’s persistent track ID. The RGB values are deterministically generated via a hashing function to ensure consistency across frames. RGB values vary within $[50, 255]$ for all three color channels. This allows the model to temporally link the same agent across timesteps and learn coherent motion patterns. The use of color fills avoids the need for explicit ID embeddings and leverages the spatial structure of the image.

\begin{wraptable}{r}{0.45\textwidth}
\centering
\vspace{-10pt}
\small
\caption{Class encoding color scheme for bounding box border color}
\begin{tabular}{llc}
  \toprule
  \textbf{Class} & \textbf{Border Color} & \textbf{RGB values} \\
  \midrule
  person      & Blue   & (0, 0, 255) \\
  car         & Red    & (255, 0, 0) \\
  truck       & Orange & (247, 162, 44) \\
  bus         & Yellow & (250, 255, 2) \\
  train       & Green  & (0, 255, 0) \\
  motorcycle  & Purple & (204, 153, 155) \\
  bicycle     & Pink   & (255, 209, 22) \\
  \bottomrule
\end{tabular}
\label{tab:class_colors}
\vspace{-10pt}
\end{wraptable}

\paragraph{Class Encoding (Border Color)}
To distinguish between different types of road users (e.g., cars, trucks, buses, pedestrians, cyclists), we draw a thin border around each bounding box in a class-specific color. These colors are chosen from a fixed palette (see Table~\ref{tab:class_colors}), and the mapping between semantic classes and RGB border values is consistent across all training data . This helps the model differentiate object behaviors by class, which is particularly useful in crash prediction (e.g., trucks tend to behave differently from bicycles).

The final control frame is an RGB image of the same resolution as the input video (e.g., $512 \times 320$) and can represent any number of agents per frame. If not depth information is available, overlapping boxes are drawn in arbitrary order, with later boxes overwriting earlier ones.

These control frames are encoded using the same VAE encoder used for the initial frame. The resulting latent tensor is passed through a ControlNet branch and injected into the U-Net backbone of the diffusion model at selected layers during training and inference. This spatial representation allows the model to attend to object motion in a dense and learnable way without requiring symbolic or token-level processing.

This conditioning mechanism supports variable numbers of agents and enables fine-grained motion control over multiple timesteps. It also integrates seamlessly into the denoising process of the diffusion model, allowing consistent agent-level motion to be expressed throughout the generated video sequence.

\clearpage
\section{FVD and JEDi Metrics Computation}
\label{appendix:metrics}

We evaluate the quality of generated videos using both distributional video-level metrics and frame-level fidelity metrics. Fréchet Video Distance (FVD)~\cite{unterthiner2019fvd} measures the distance between the distributions of generated and real videos (lower is better). This is done by embedding the two distributions in the feature space of a pretrained Inflated 3D ConvNet (I3D) and computing the Fréchet distance under the assumption of a multivariate Gaussian. JEDi~\cite{luo2025jedi} addresses limitations of FVD by relaxing the Gaussian assumption, employing features from a Joint Embedding Predictive Architecture (JEPA), and using Maximum Mean Discrepancy (MMD) with a polynomial kernel to improve both temporal sensitivity and sample efficiency. Together, these metrics provide complementary views of video realism at the distributional level. 


To evaluate distributional similarity between generated and real videos, we compute Fréchet Video Distance (FVD) and JEDi under two complementary evaluation protocols, which differ based on how the ground-truth distribution is constructed:

\begin{enumerate}
    \item \textbf{Condition-Aligned Evaluation:}
    For models such as Ctrl-Crash, Ctrl-V, and SVD Base, we have access to the ground-truth data used to seed generation—namely, the initial frame, bounding box trajectories, and even crash type category. In this setting, we sample 200 videos from the MM-AU validation set, and generate corresponding videos using their conditioning information. We do not directly compare generated videos to their exact ground-truth counterpart; instead, we treat these 200 ground-truth videos as a reference distribution that is semantically aligned with the generated one (e.g., similar vehicle types, scene layouts, and crash categories). FVD and JEDi are then computed between these two condition-aligned distributions. This setup ensures that the real and generated samples share similar structural priors, making the distributional comparison more meaningful. This method is used for FVD and JEDi results reported in Table~2 and Table~3 of the main paper.

    \item \textbf{Random GT Evaluation:}  
    Some baselines, such as AVD2, only provide generated videos without access to the original conditioning inputs, preventing us from constructing a semantically aligned ground-truth distribution. To ensure fair comparison in such cases, we instead sample 500 videos randomly from the MM-AU validation set to represent the ground-truth distribution. We then compare this to 200 generated samples from each model (including AVD2, Ctrl-Crash, Ctrl-V, and SVD Base). While this evaluation provides less precise alignment between real and generated content, it allows for inclusion of baselines where seed information is unavailable. This method is used for FVD and JEDi results reported in Table~1 of the main paper.
\end{enumerate}

By using both evaluation protocols, we balance semantic alignment (when possible) with broad comparability, enabling fair and informative assessments across different classes of models.

In addition, we report frame-level metrics that directly assess perceptual quality relative to ground truth frames. These include LPIPS (Learned Perceptual Image Patch Similarity)~\cite{zhang2018lpips}, SSIM (Structural Similarity Index), and PSNR (Peak Signal-to-Noise Ratio). These metrics quantify visual similarity, structural consistency, and reconstruction fidelity, respectively, offering finer-grained insight into how well the generated frames preserve appearance and local coherence. To extend these frame-level metrics to video, we average the scores for all the frames of a given video.

\clearpage
\section{Training and Implementation Details}
\label{appendix:training}

\paragraph{Model Architecture} Ctrl-Crash builds on the Stable Video Diffusion (SVD) framework as the base image-to-video generation model. In the first stage, we fine-tune the SVD model on curated crash and driving clips from the MM-AU dataset. In the second stage, we freeze the base model and train a ControlNet module to integrate spatial (bounding box control frames) and semantic (crash type) inputs via additional encoder and cross-attention layers. All control signals are processed using the same pretrained VAE encoder used by SVD. The number of parameters for each model (and each sub-module) is given in Table~\ref{tab:num_parameters}.

\paragraph{Training Setup} We use the AdamW optimizer with a constant learning rate of $4\times10^{-5}$ and batch size of $1$. Extensive hyperparameter tuning was intentionally avoided, as our focus is on validating the method’s effectiveness rather than maximizing performance through optimization. The first-stage fine-tuning of the base SVD model is performed for $101$k steps, using an MSE loss in latent space. The second-stage ControlNet training is run for $31$k steps, with conditioning dropout applied as described in Section~3.4 (main paper). We use mixed precision during training by setting weights and inputs to fp16 for non-trainable (frozen) parts of the model (i.e., VAE encoder, VAE decoder, CLIP encoder) and keep the trainable parts at fp32 to reduce memory usage. Training is performed on 4 NVIDIA 80GB A100 GPUs over approximately 2 weeks for both stages combined. All models are implemented in PyTorch using the Hugging Face accelerate library as a base.

\paragraph{Inference and Guidance Parameters} During inference, we apply multi-condition classifier-free guidance with tunable guidance scales. Unless otherwise specified, we use $\gamma_B \in [1, 3] \subset \mathbb{R}$ for bounding box control and $\gamma_T \in [6, 12] \subset \mathbb{R}$ for crash type conditioning. These values represent the ranges of values within the guidance scales will increase linearly throughout the denoising process (e.g., guidance scale starting at 1 at first denoising step and finishes at 3 at the last denoising step). The base model $\epsilon_\phi$ used for unconditional noise prediction is the original pretrained SVD base checkpoint prior to any fine-tuning. We sample videos with 25 frames at a resolution of $512 \times 320$, using DDIM sampling with $30$ denoising steps.

\paragraph{Dataset Splits} We curate a training set of $7,500$ clips from MM-AU after filtering. A held-out validation set of $900$ clips is used for evaluation. All reported metrics in the main paper (Tables~1, 2) are computed on generated samples from the held-out validation set. 

\begin{table}[htbp]
    \centering
    \vspace{5mm}
    \begin{tabular}{lccc}
    \toprule
    \textbf{Submodule} & \textbf{Status (Stage 1)} & \textbf{Status (Stage 2)} & \textbf{Number of Parameters}\\
    \midrule
    VAE-Encoder & Frozen & Frozen & 34,163,592 \\
    VAE-Decoder & Frozen & Frozen & 63,579,183\\
    CLIP-Image Encoder & Frozen & Frozen & 632,076,800 \\
    UNet & Trainable & Frozen & 1,524,623,082\\
    ControlNet & N/A & Trainable & 681,221,585\\
    \midrule
    Total & & & 2,935,664,242 $\approx$ \textbf{3B} \\
    \bottomrule
    \end{tabular}
    \vspace{5mm}
     \caption{Number of parameters by submodule. Refer to architecture diagram in Figure~2 of the main paper for more information on the submodules. Stage 1 and Stage 2 refer to the two stages of training for our method.}
    \label{tab:num_parameters}
\end{table}

\clearpage
\section{Dataset}
\label{appendix:datasets}

Dataset annotations will be made publicly available and it will be possible to obtain them by following instructions from our open-source code base. We include a subset of the annotations in the supplementary ZIP file for reference, as the full set would exceed the size limit for submission.

All annotated video samples in our dataset (MM-AU extension, RussiaCrash test set, etc.) are stored in JSON format, where each annotation file corresponds to a single video. Below, we describe the structure and semantics of the annotation schema.

\begin{figure}[htbp]
  \centering
  \makebox[\textwidth][c]{\includegraphics[width=\linewidth]{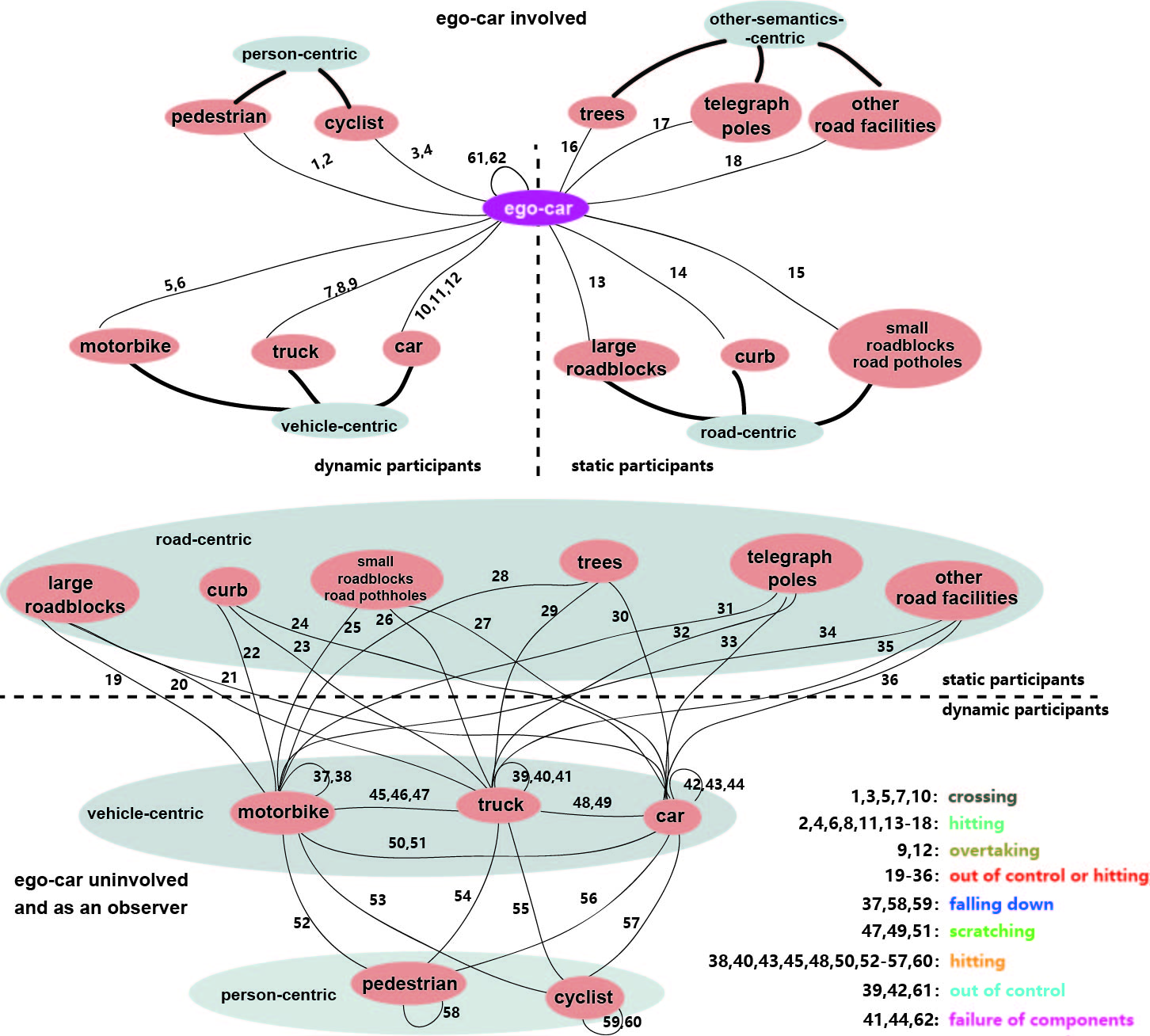}}
  \caption{Original accident type definition used by the MM-AU dataset~\cite{fang2024abductive} as defined by the DADA2000 dataset~\cite{fang2023dadadriverattentionprediction}}
  \label{fig:dada_crashtypes}
\end{figure}

\paragraph{Annotation Structure} Each annotation consists of three main fields:

\begin{itemize}
    \item \texttt{video\_source}: The filename of the source video (e.g., \texttt{"7\_00951.mp4"}).
    
    \item \texttt{metadata}: High-level information about the annotated scenario, including:
    \begin{itemize}
        \item \texttt{ego\_involved} (bool): Indicates whether the ego vehicle (assumed to be the camera holder) is involved in the accident.
        \item \texttt{accident\_type} (int): A categorical index representing the accident type as defined by DADA2000 dataset. See Figure~\ref{fig:dada_crashtypes} and Table~\ref{tab:class_count} for more information).
    \end{itemize}
    
    \item \texttt{data}: A list of per-frame annotations. Each frame entry includes:
    \begin{itemize}
        \item \texttt{image\_source}: The filename of the corresponding frame image (e.g., \texttt{"7\_00951\_0000.jpg"}).
        \item \texttt{labels}: A list of annotated objects (bounding boxes) in the frame. Each object contains:
        \begin{itemize}
            \item \texttt{track\_id} (int): A persistent ID assigned to each object across frames.
            \item \texttt{name} (str): The object class as a string (e.g., \texttt{"car"}, \texttt{"person"}, \texttt{"truck"}).
            \item \texttt{class} (int): The numerical class index used internally (e.g., 0 = person, 1 = car).
            \item \texttt{box} (list[float]): The bounding box coordinates, normalized to the range [0, 1], in the format \texttt{[x\_min, y\_min, x\_max, y\_max]}.
        \end{itemize}
    \end{itemize}
\end{itemize}

All bounding boxes are temporally linked using consistent \texttt{track\_id} values. Object classes follow the taxonomy defined by the YOLOv8 model used in the annotation pipeline.

The MM-AU dataset~\cite{fang2024abductive} labels each video with an accident type as defined by the DADA2000 dataset~\cite{fang2023dadadriverattentionprediction}. These accident types are presented in Figure~\ref{fig:dada_crashtypes}. In our work, we reduce these numerous crash types to 5 types as defined in Section~3.3 and repeated here: Crash types are represented by
a discrete class label from five categories: (0) none, (1) ego-only, (2) ego/vehicle, (3) vehicle-only,
and (4) vehicle/vehicle. These indicate which agents are involved in the crash, with, for example,
“vehicle/vehicle” describing a crash between two non-ego agents. Table~\ref{tab:class_count} shows the association from the DADA2000 crash types to the Ctrl-Crash crash types that were used in this work.

\begin{table}[htb]
  \caption{Ctrl-Crash to DADA2000 crash type association}
  \centering
  \begin{tabular}{lcc}
    \toprule
    \textbf{Crash Type} 
    & \textbf{DADA2000 Crash Types}
    & \textbf{\# of Training Samples} \\
    \midrule
    0 - no crash & N/A & $1745$ \\
    1 - ego-only & 13, 14, 15, 16, 17, 18, 61, 62 & $267$ \\
    2 - ego/vehicle & 1-12 & $3182$ \\
    3 - vehicle-only & 19-37, 39, 41, 42, 44 & $577$ \\
    4 - vehicle/vehicle & 38, 40, 43, 45-51 & $2168$ \\
    \bottomrule
  \end{tabular}
  \label{tab:class_count}
\end{table}

For training and evaluation, we split all videos into 25-frame clips. The MM-AU dataset gives exact frame labeling to indicate when the accident occurs and when abnormal driving starts and ends before and after the accident. For each video we sample a 25-frame video containing the accident frame and label it according to the accident type. Then we sample, if possible, a 25-frame video containing only "normal" driving (i.e., with no overlap with abnormal or accident labeled frames) and label these clips with a crash type "no crash". Ultimately this yields $6,927$ clips containing a crash and $1,964$ clips without a crash (see Table~\ref{tab:class_count} for number of clips for each class during training).

\clearpage
\section{User Survey}
\label{appendix:survey}

\begin{figure*}[!htb]
    \begin{center}
        
    \begin{subfigure}{0.45\textwidth}
        \includegraphics[width=0.8\linewidth]{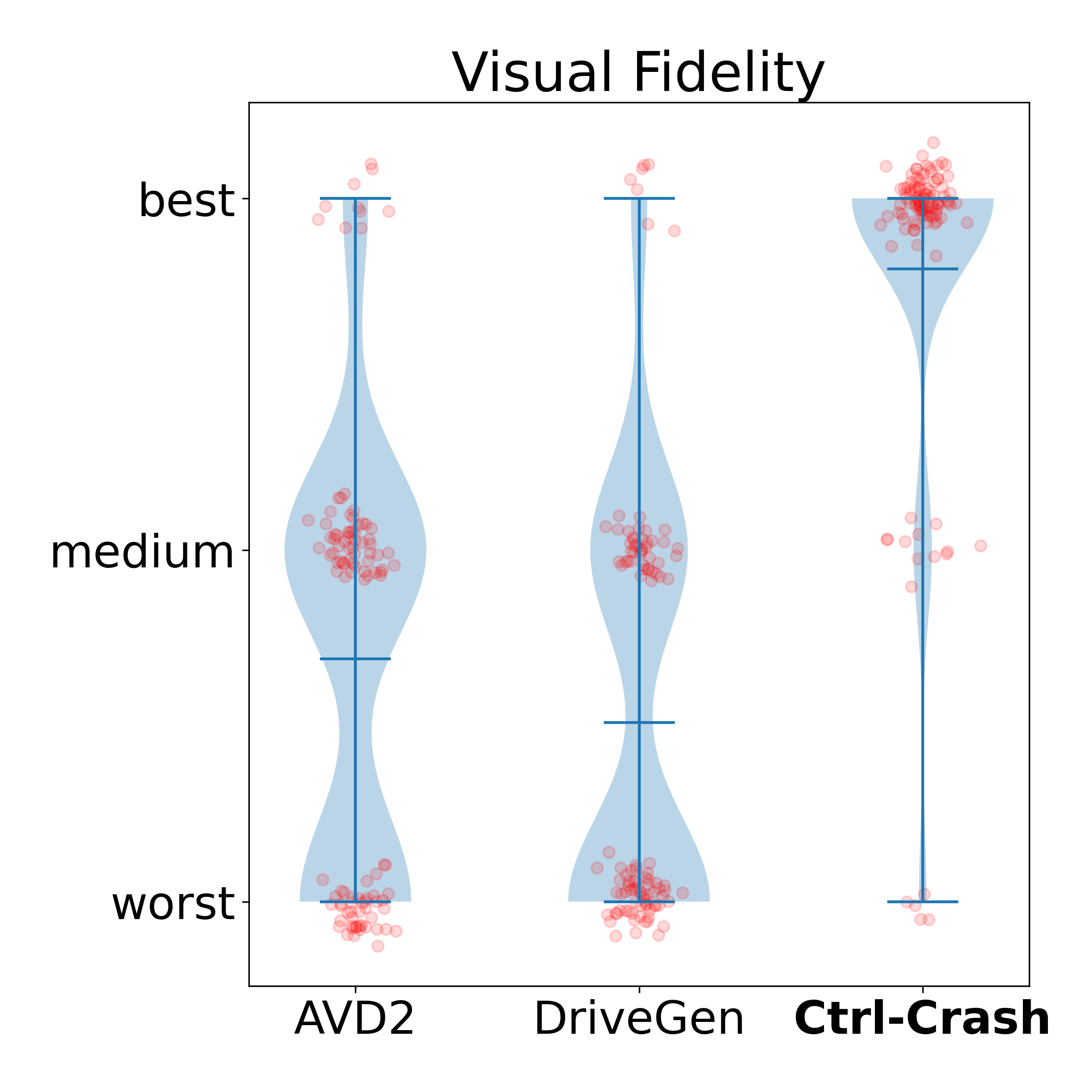} 
        \caption{Visual preference comparison in generated videos.}
        \label{fig:subim1}
    \end{subfigure}\hspace{10mm}
    \begin{subfigure}{0.45\textwidth}
        \includegraphics[width=0.8\linewidth]{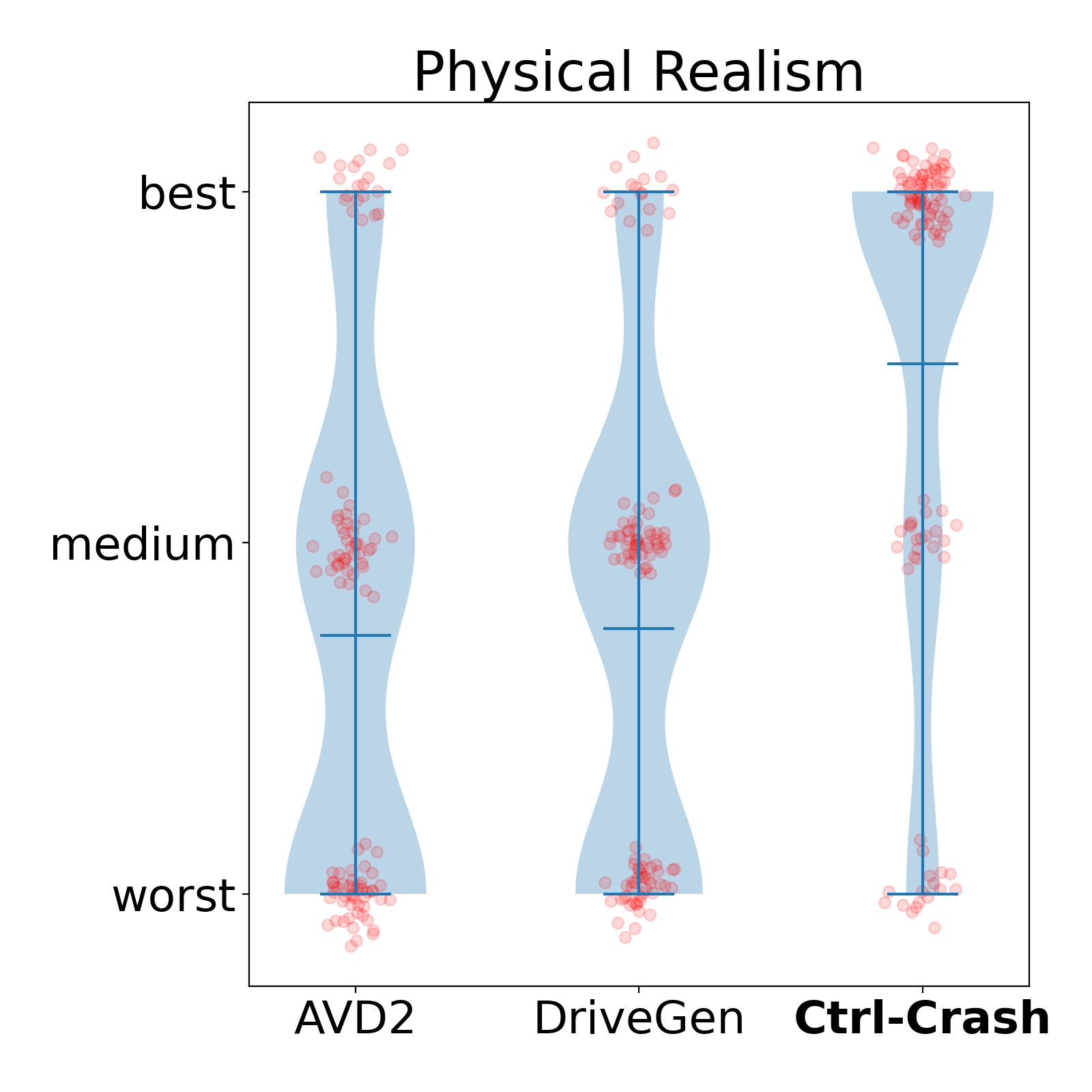}
        \caption{Realism preference comparison in physical appearance.}
        \label{fig:subim2}
    \end{subfigure}
    \end{center}
    
    \caption{User study results comparing generated videos from AVD2, DrivingGen and Ctrl-Crash with 40 participants across 5 crash types: Participants exhibit a strong preference for Ctrl-Crash generated videos, citing superior visual quality and physical realism.}
    \label{fig:user_study}
\end{figure*}

We conducted a brief user study with $n=21$ participants who were asked to rank $k=3$ videos (from Ctrl-Crash, AVD2, and DrivingGen) across 5 different crash scenarios each. 
The participants consist of students from our lab who are not associated with the project in any way.
No further demographic data was collected.
The users were asked to rank the 3 videos in each of the 5 situations by (a) physical plausibility and (b) visual fidelity from best to worst.
The users were required to choose a best/medium/worst video in each question and visual/physical category.
We used the non-parametric Friedman test \cite{friedman1937use} to determine with $p\leq0.01$ that there is a method that is consistently ranked higher than the others (Ctrl-Crash). 
We further used the Nemenyi post-hoc analysis \cite{nemenyi1963distribution} to find that with $p\leq0.01$, our method consistently outperforms both AVD2 and DrivingGen in both physical realism and visual fidelity (see Figure~\ref{fig:user_study}).

The participants were not given any information about the study ahead of time other than that it revolves around state-of-the-art video generation.
The survey was carried out through Google Forms.
At the start of the survey, they received the following instructions:

\begin{displayquote}
\textbf{Content warning:} ai-generated mild car crashes with no humans depicted. No blood/injury, just car-on-car action.\\

Please help us get some human feedback on a new video generation method.
We're asking you to rank 5 sets of videos. 
Should take less than 5 min, and you'll see why AI won't take over anything anytime soon. 

\textbf{Instructions:}
For each of these 5 questions, we will show you 3 short video clips that have ALL been AI-generated.
We will ask you:
\begin{itemize}
    \item Does each video depict a crash?
    \item Rank the 3 videos by highest physical accuracy/plausibility
    \item Rank the 3 videos by highest visual fidelity (aka are they nice to look at)\\
\end{itemize}

\textbf{Important:}
Please rank the videos relative to one another, i.e. "best" means best of the 3.
\end{displayquote}

\begin{landscape}
\begin{figure}[htbp]
  \centering
  \begin{minipage}[b]{0.42\linewidth}
    \centering
    \includegraphics[width=\linewidth]{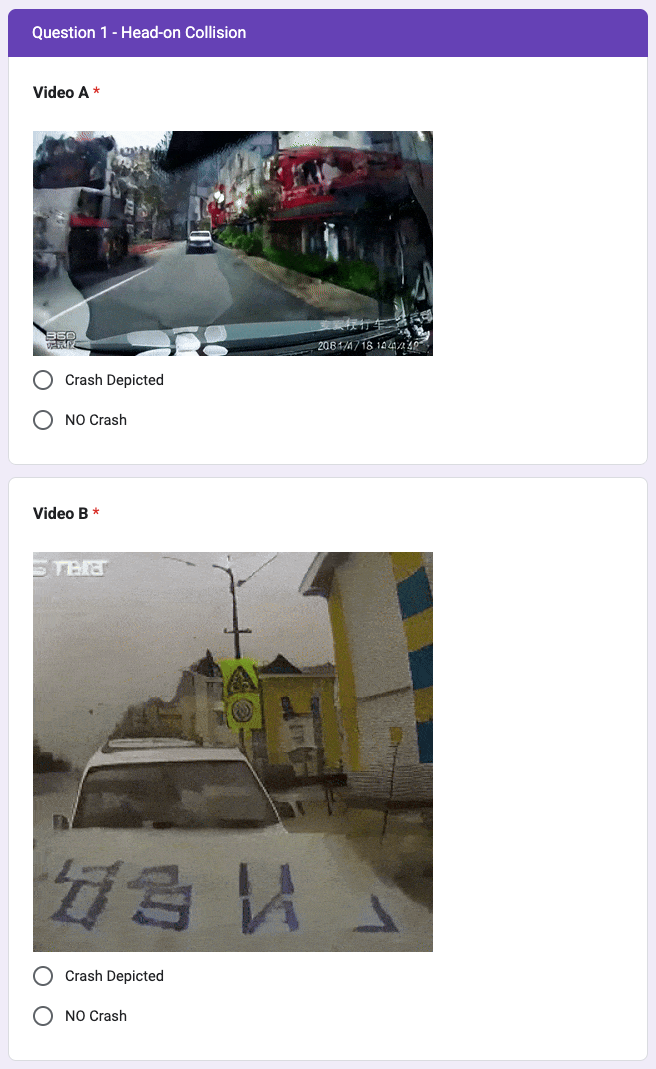}
    \caption{User Survey Screenshot 1, showing 2 (out of 3) samples from question 1. What is shown as static image here was a GIF in the original Google Form.}
    \label{fig:gform-screen1}
  \end{minipage}%
  \hfill
  \begin{minipage}[b]{0.42\linewidth}
    \centering
    \includegraphics[width=\linewidth]{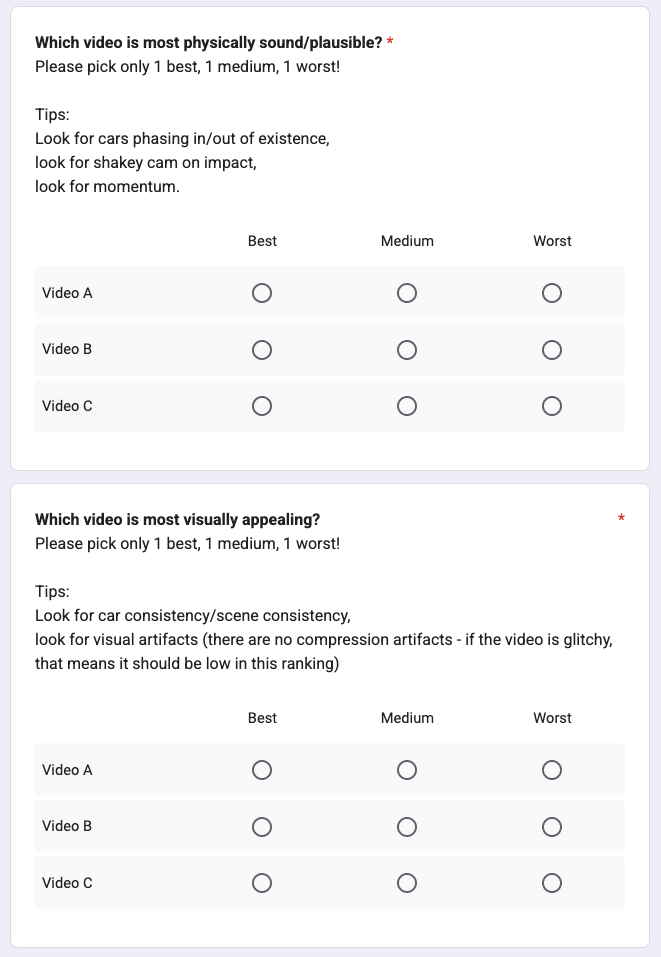}
    \caption{User Survey Screenshot 2, showing the evaluation questions for each batch of 3 videos. Users were forced to rank all videos and to only use each rank once (i.e., it is not possible to submit the form when more than one video in each batch has the same rank).}
    \label{fig:gform-screen2}
  \end{minipage}
\end{figure}
\end{landscape}

Each of the 5 accident types (with 3 videos each, labeled A,B,C, randomly shuffled from AVD2, DriveGen, ours) were presented like in Figure~\ref{fig:gform-screen1}. The 5 accident types we selected were: head-on collision, t-bone, rear-ending, dangerous overtaking, and loosing control of the vehivle and going off the road.
The ranking was implemented as shown in Figure~\ref{fig:gform-screen2} by forced-choice.

\end{document}